\documentclass[pdflatex,sn-nature]{sn-jnl}
\usepackage{graphicx}
\usepackage{amsmath,amssymb,amsfonts,bm}
\usepackage{amsthm}
\usepackage{upgreek}
\usepackage{booktabs}
\usepackage[title]{appendix}
\usepackage{mathrsfs}
\usepackage{xcolor}
\usepackage{textcomp}
\usepackage{manyfoot}
\usepackage{enumitem}
\usepackage{verbatim}
\usepackage{epsfig}
\usepackage{epstopdf}
\usepackage{helvet}
\usepackage{courier}
\usepackage[bottom]{footmisc}
\usepackage{longtable}
\usepackage{flafter}
\usepackage{placeins}
\usepackage{needspace}
\usepackage{geometry}
\graphicspath{{picture/}}
\begin{document}
\title[Online Active Learning Architecture]{Online Active Learning Architecture for Safety-Critical Control Systems}

\author[1,5,*]{Shuiwen Shen}
\author[2]{Jinhua Zhang}
\author[3]{Yan Wang}
\author[4]{Huanqiang Zeng}

\makeatletter
\g@addto@macro\artauthors{%
	\par\noindent
	\def\authorsep{}%
}
\makeatother

\author[2]{Yueyun Xu}
\author[2]{Song Gao}
\author[1]{Haifeng Xiang}
\author[1]{Yifu Shen}

\affil[1]{School of Mechanical and Automotive Engineering, Xiamen University of Technology, Xiamen, 361024, China}
\affil[2]{China Innovation Center for Intelligent and Connected Vehicles, Beijing, 100176, China}
\affil[3]{SAMR Defective Product Recall Technical Center (Motor Vehicle Institute), Beijing, 100101, China}
\affil[4]{School of Opto-Electronic and Communication Engineering, Xiamen University of Technology, Xiamen, 361024, China}
\affil[5]{State Innovative Centre for Automotive Electric Driving at Xiamen, Xiamen, Fujian, 361010, China}
\affil[*]{2023000085@xmut.edu.cn}

\abstract{The transition of information-centric artificial intelligence to safety-critical applications is impeded by the limitations of conventional offline-training paradigms in open-world environments. This paper presents Control AI, an architecture for Online Active Learning (OAL) inspired by the continuous Prediction-Evaluation-Correction process of biological intelligence. The framework combines the hierarchical representation capacity of deep neural networks with the real-time error-feedback adaptation of adaptive control. A prediction model maps sensory input into a latent manifold, an optional latent model represents transitions in that space, and a reconstruction model maps the latent state back to observable signals. Residuals between predictions and measurements provide the internal learning signal. Spatial and temporal cross-reconstruction architectures prevent OAL from collapsing to a trivial identity mapping, while coupled, dual, and localized variants shorten error-transmission paths. Localized predictive coding style OAL is implemented through guidance-backpropagation and target-backpropagation architectures, and hybrid and scaled variants support high-dimensional deployment. The paper focuses on the architectural rationale and the progression from a conceptual feedback loop to practical large-scale configurations.}
\maketitle
\thispagestyle{empty}

\section{Introduction}\label{introduction}

The rapid development of artificial intelligence (AI) has largely been driven by the mastery of data and information. Large language models such as ChatGPT and Gemini have transformed code generation and natural-language processing \cite{Naddaf2023ChatGPT,team2023gemini}, while deep vision systems perform powerful probabilistic recognition within massive but bounded data distributions. This paradigm is effective for informational AI, whose principal objective is to infer statistical relationships from a predefined corpus. It is insufficient, however, when AI is embedded in a physical system. Safety-critical control requires an agent not only to infer, but also to remain stable and adapt while interacting with an open, non-stationary, and non-cooperative environment. Control AI is therefore needed as a closed-loop complement to informational AI rather than as a larger version of the same offline-trained model.

The distinction becomes concrete in autonomous driving, where long-tail conditions persist in unconstrained physical environments \cite{Ananthaswamy2024PBrain,wang2024survey}. Offline-trained models are normally deployed with fixed parameters and must interpret novel states through priors formed before deployment. The October 2023 Cruise incident and recurrent collisions with stationary emergency vehicles show how departures from the training corpus can become unsafe physical behavior when errors enter a control loop \cite{NHTSA2024Cruise,koopman2017autonomous,wansley2023regulating,NHTSA2023Report}. Data scaling alone cannot remove this structural mismatch.

The mathematical source lies in the learning mechanism. Conventional deep learning transmits a predefined global loss through the full layer chain by backpropagation \cite{Rumelhart1986BackProp,LeCun2015DeepLearning,lillicrap2020backpropagation}. Although effective for hierarchical representation, it separates global offline training from largely fixed online inference and provides no intrinsic local signal relating a current prediction to the current physical state. Long Jacobian products also hinder continuous high-dimensional adaptation. Biological intelligence instead couples local, continuous learning to perception \cite{Hinton2022ForwardForward,richards2019deep,friston2009theory,kudithipudi2022biological}. Predictive coding generates an endogenous prediction, compares it with sensory evidence, and uses the residual for correction \cite{friston2010Brain,zador2019critique,knill2004bayesian,rao1999predictive,Hosoya2005PC}. This Prediction-Evaluation-Correction cycle replaces a remote global objective with locally available discrepancies \cite{Gabhart2025PC,ccatal2020learning}.

This paper translates that mechanism into an Online Active Learning (OAL) architecture for Control AI. A prediction model maps sensory input into a latent manifold, an optional latent model represents its evolution, and a reconstruction model returns the latent state to the observable signal. The residual between reconstruction and measurement closes a physically grounded, self-supervised learning loop, allowing hierarchical error modules to update parameters near the representation that generated the discrepancy. Spatial cross-reconstruction and temporal prediction prevent trivial identity mapping; coupled and dual variants shorten error paths; localized predictive-coding implementations provide guidance- and target-backpropagation; and hybrid and scaled variants distribute the loops across large networks.

The resulting architecture preserves the hierarchy of informational AI while adding the real-time plasticity required by Control AI. A high prediction residual indicates a mismatch between endogenous expectations and sensory observations and can therefore signal an out-of-distribution change. Localized learning and damping mechanisms support continuous adaptation while maintaining stable online learning dynamics. This paper focuses on the architectural mechanism and on the progression from the conceptual feedback loop to practical high-dimensional configurations. The complete conditions under which interacting local loops yield global stability are developed in a separate companion stability paper, and task-specific validation is reserved for subsequent studies.

Section 2 compares deep and adaptive learning, constructs the OAL feedback mechanism, derives the prediction-latent-reconstruction architecture, and develops spatial, temporal, coupled, dual, localized predictive-coding, hybrid, and scaled implementations. Section 3 summarizes the contribution and its relationship to the companion stability analysis.

\FloatBarrier
\section{Online Active Learning Architecture}\label{online-active-learning-architecture}

\FloatBarrier
\subsection{Deep Learning and Adaptive Learning}\label{deep-learning-and-adaptive-learning}

\begin{figure}[!h]
	\begin{center}\footnotesize
		\begin{tabular}{l@{\qquad}l}
			\raisebox{-0.5\height}{\includegraphics[scale=0.36]{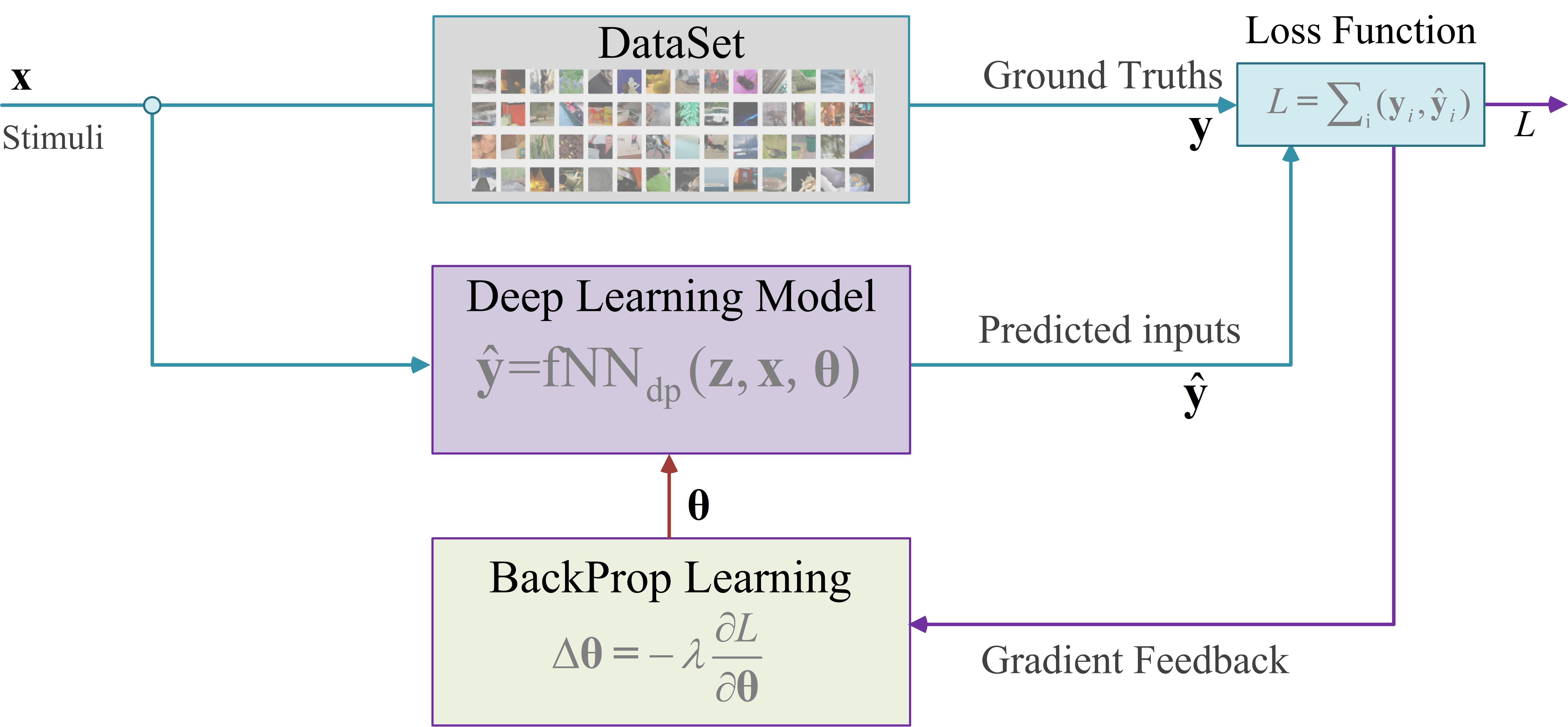}} &
			\raisebox{-0.5\height}{\includegraphics[scale=0.36]{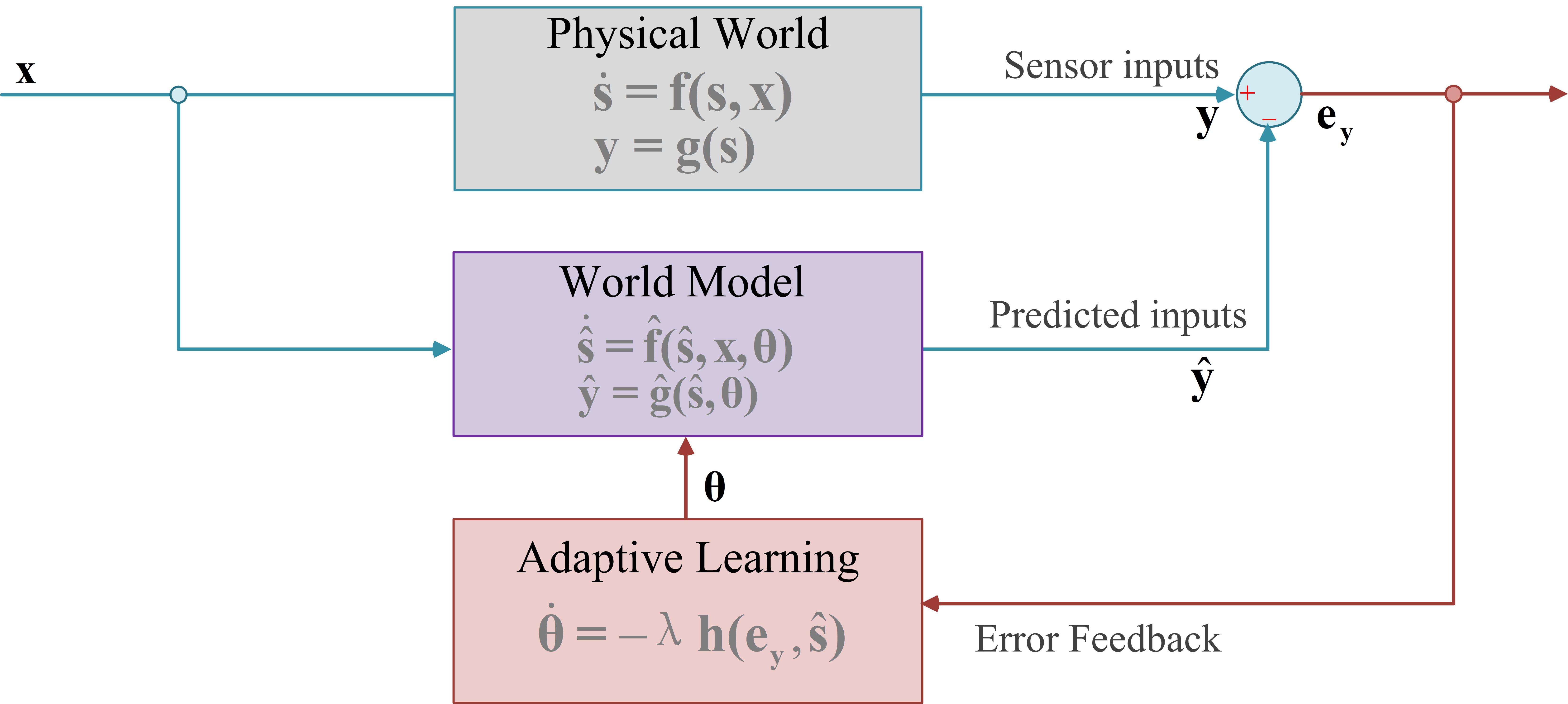}}\\
			\addlinespace[1.5ex]
			a) Deep Back-Propagation Offline Training ($\mathbf{F_{BP}}$) & b) Adaptive Online Learning ($\mathbf{F_{AP}}$)\\
			\addlinespace[3ex]
			\raisebox{-0.5\height}{\includegraphics[scale=0.36]{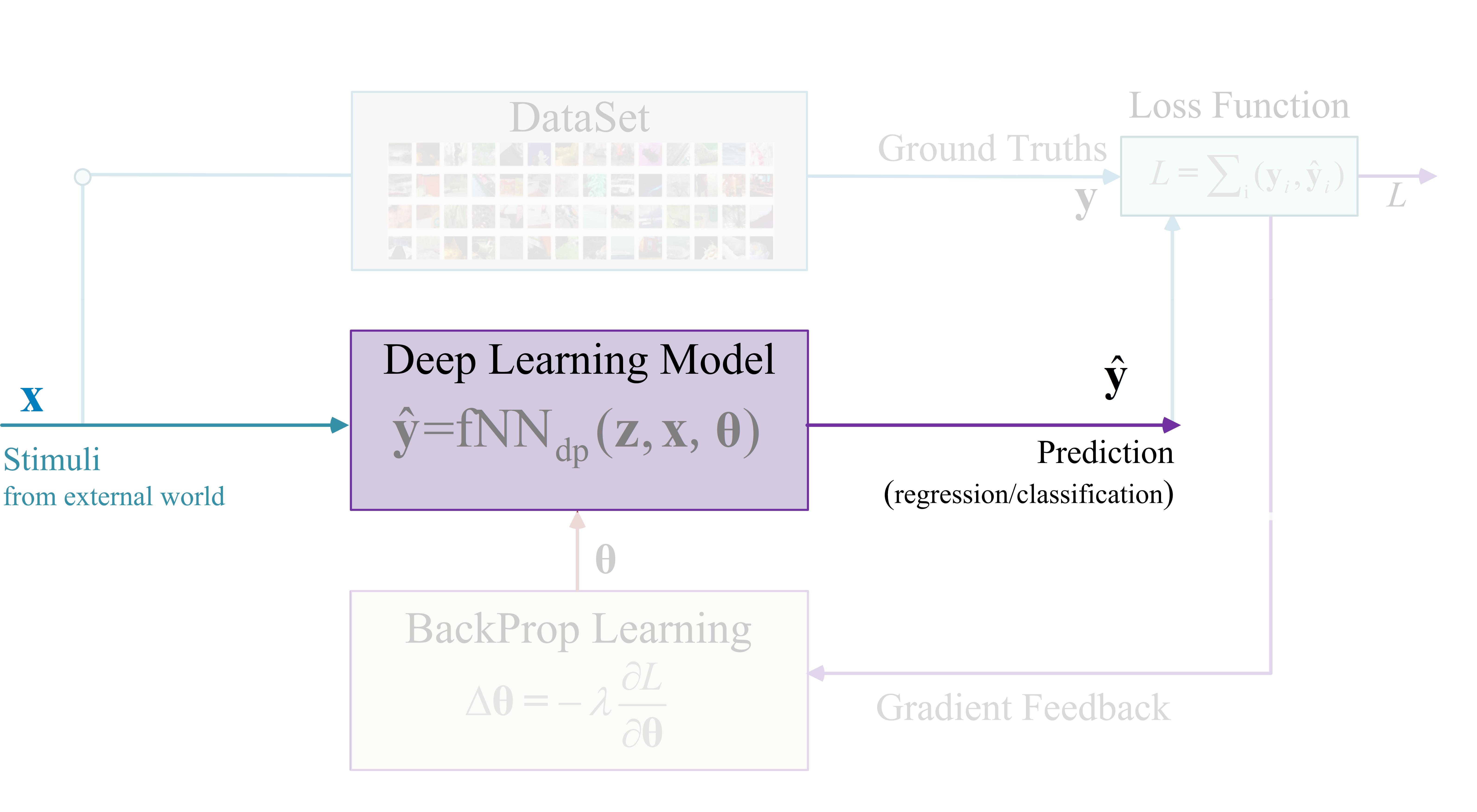}} &
			\raisebox{-0.5\height}{\includegraphics[scale=0.36]{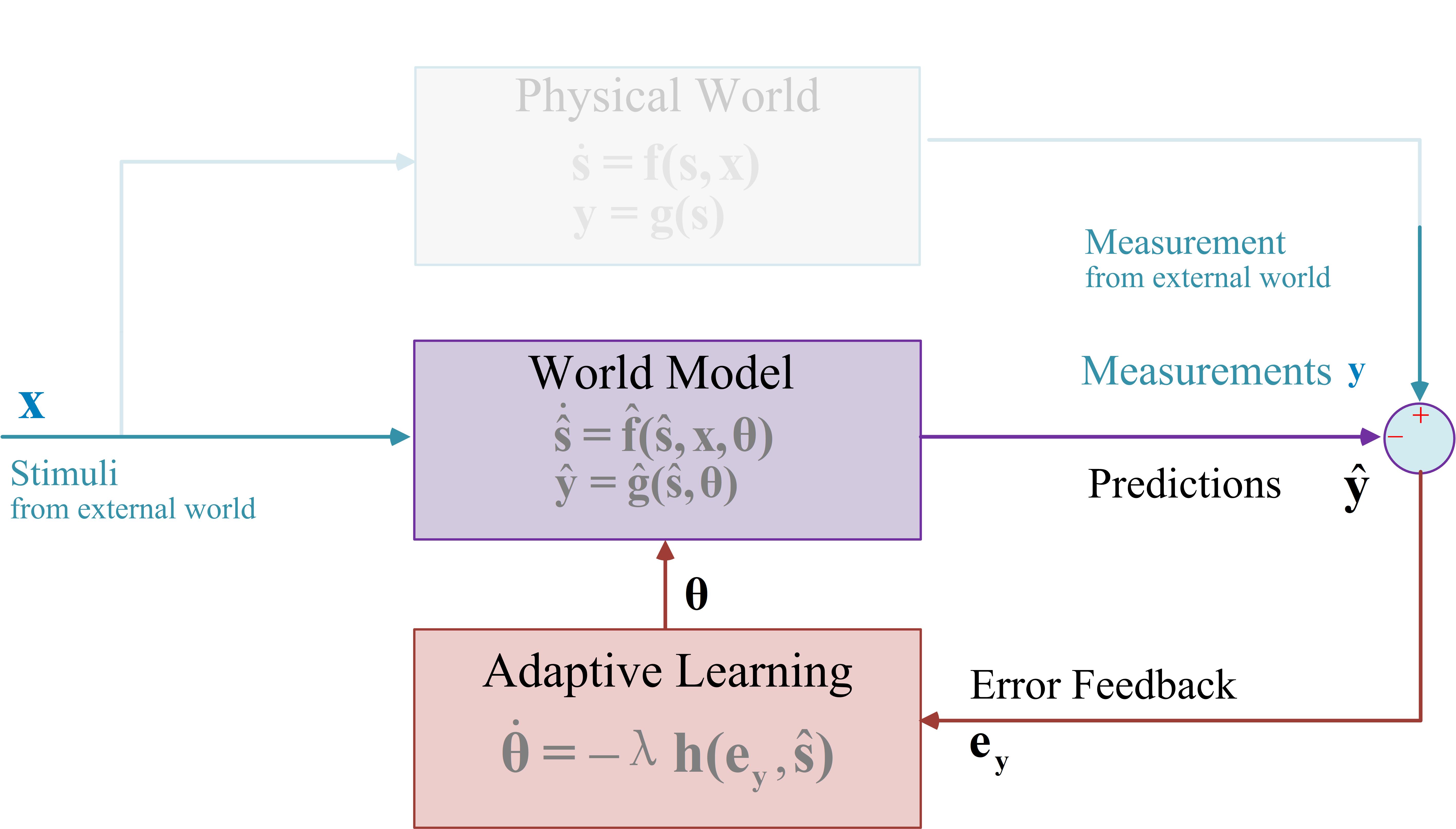}}\\
			\addlinespace[1.5ex]
			c) Deep Forward Online Inferencing & d) Adaptive Online Inferencing\\
			\addlinespace[3ex]
			\raisebox{-0.5\height}{\includegraphics[scale=0.36]{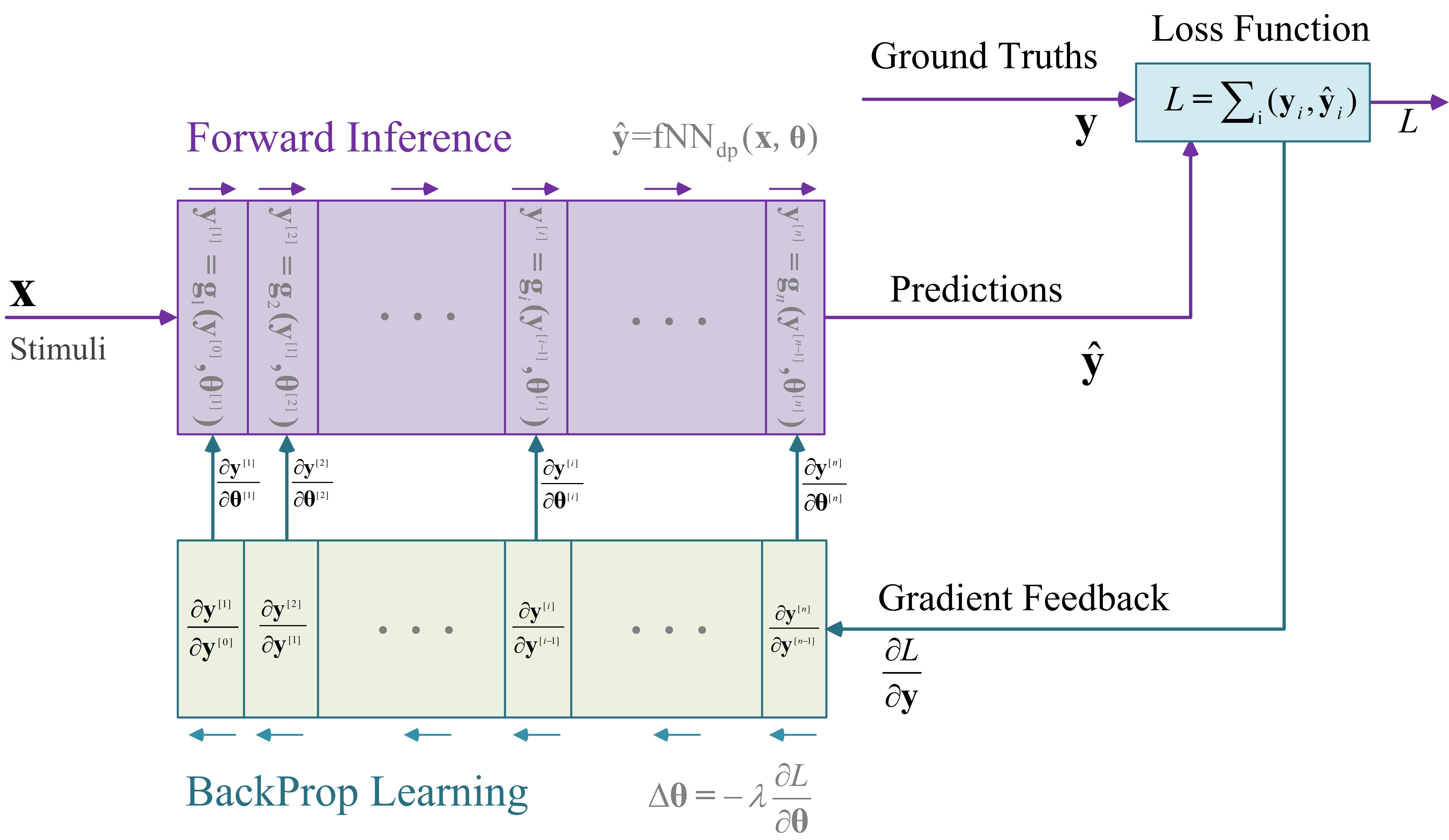}} &
			\raisebox{-0.5\height}{\includegraphics[scale=0.36]{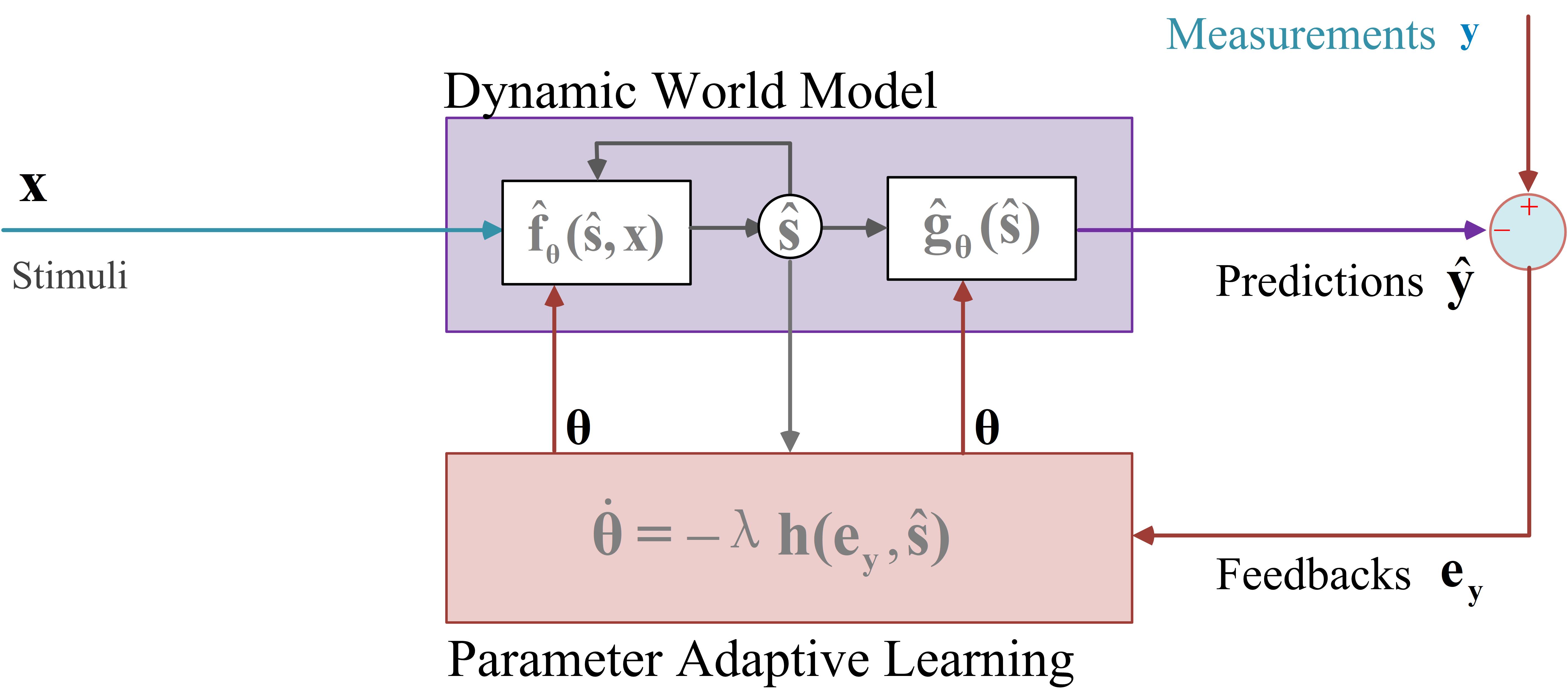}}\\
			\addlinespace[1.5ex]
			e) Deep Structured Training & f) Adaptive Un-structured Learning
		\end{tabular}
	\end{center}
	\vspace{0.5em}
	\caption{Deep learning vs Adaptive Learning}
	\label{fig:paradigms}
\end{figure}

Despite substantial architectural differences among multilayer perceptrons, convolutional neural networks, recurrent neural networks, Transformers, and their hybrids, most deep models for prediction and object detection share a common structural framework. Fig.~\ref{fig:paradigms}(a), (c), and (e) show the corresponding offline-training, online-inference, and structured-gradient paths.

A deep-learning model takes a vector \(\mathbf{x}\), such as image pixels, as its input and produces a prediction \(\hat{\mathbf{y}}\) using pretrained parameters \(\bm{\uptheta}\):

\begin{equation}
\hat{\mathbf{y}}={f\mathbb{NN}}_\mathrm{dl}(\mathbf{x},\bm{\uptheta})
\end{equation}

Modern models may contain billions of parameters. A defining property is their hierarchical organization:

\begin{equation}
\begin{array}{rcl}
\mathbf{z}^{[i]} & = & \bm{\uptheta}^{[i,i-1]}\mathbf{y}^{[i-1]} \\
\mathbf{y}^{[i]} & = & g_i(\mathbf{z}^{[i]})
\end{array}
\end{equation}

The latent states \(\mathbf{z}^{[1]},\mathbf{z}^{[2]},\ldots,\mathbf{z}^{[n]}\) and intermediate outputs \(\mathbf{y}^{[1]},\mathbf{y}^{[2]},\ldots,\mathbf{y}^{[n]}\) form low-dimensional manifolds in a high-dimensional dense space. With sufficient depth and width, a deep neural network can approximate a broad class of input-output functions \cite{zhou2020universality}. The dependencies of these manifolds follow a Markov-chain information-processing principle: the representation at the current layer depends on the preceding layer rather than on later layers \cite{amjad2019learning,wu2018deep}. The forward manifolds are governed by learnable parameters \(\bm{\uptheta}^{[i,i-1]}\), abbreviated as \(\bm{\uptheta}^{[i]}\) where appropriate, and nonlinear activation functions \(g_i\).

Because the layer structures are repeated, the gradients used for offline learning follow the same hierarchy in the reverse direction:

\begin{equation}
\begin{array}{rcl}
\textbf{g}_{i_{\;\;}} & = & \frac{\partial \mathbf{y}^{[i+1]}}{\partial \mathbf{y}^{[i]}}\textbf{g}_{i+1} \\
\textbf{g}_{\bm{\uptheta}_i} & = & \frac{\partial \mathbf{y}^{[i]}}{\partial \bm{\uptheta}^{[i]}}\textbf{g}_i \\
\Delta\bm{\uptheta}^{[i]} & = & -\lambda\textbf{g}_{\bm{\uptheta}_i}
\end{array}
\end{equation}

Here \(\textbf{g}_i\) and \(\textbf{g}_{\bm{\uptheta}_i}\) are the gradients of \(\mathbf{y}^{[i]}\) and \(\bm{\uptheta}^{[i]}\) with respect to a predefined loss \(\mathcal{L}\). For classification, the loss is often defined through cross entropy,

\begin{equation}
\mathcal{L}=\textstyle\sum\nolimits_k y_k\ln p(\hat{y}_k)
\end{equation}

where \(k\) denotes the object class, \(\mathbf{y}\) or \(\mathbf{y}_{\mathrm{GT}}\) is the labeled target, and \(\hat{\mathbf{y}}\) is the prediction. The forward prediction and backward gradient paths are shown in Fig.~\ref{fig:paradigms}(e) and in the offline deep-learning frame of Fig.~\ref{fig:conceptual}.

\begin{figure}[!htbp]
	\begin{center}\footnotesize
		\begin{tabular}{c}
			\includegraphics[width=.82\textwidth]{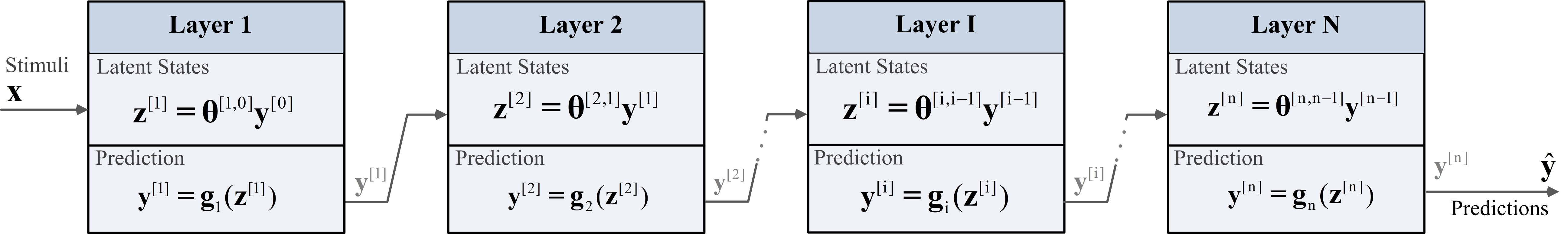}\\[0.8ex]
			a) Deep Learning Structured Online Inference Flow\\[2.0ex]
			\includegraphics[width=.78\textwidth]{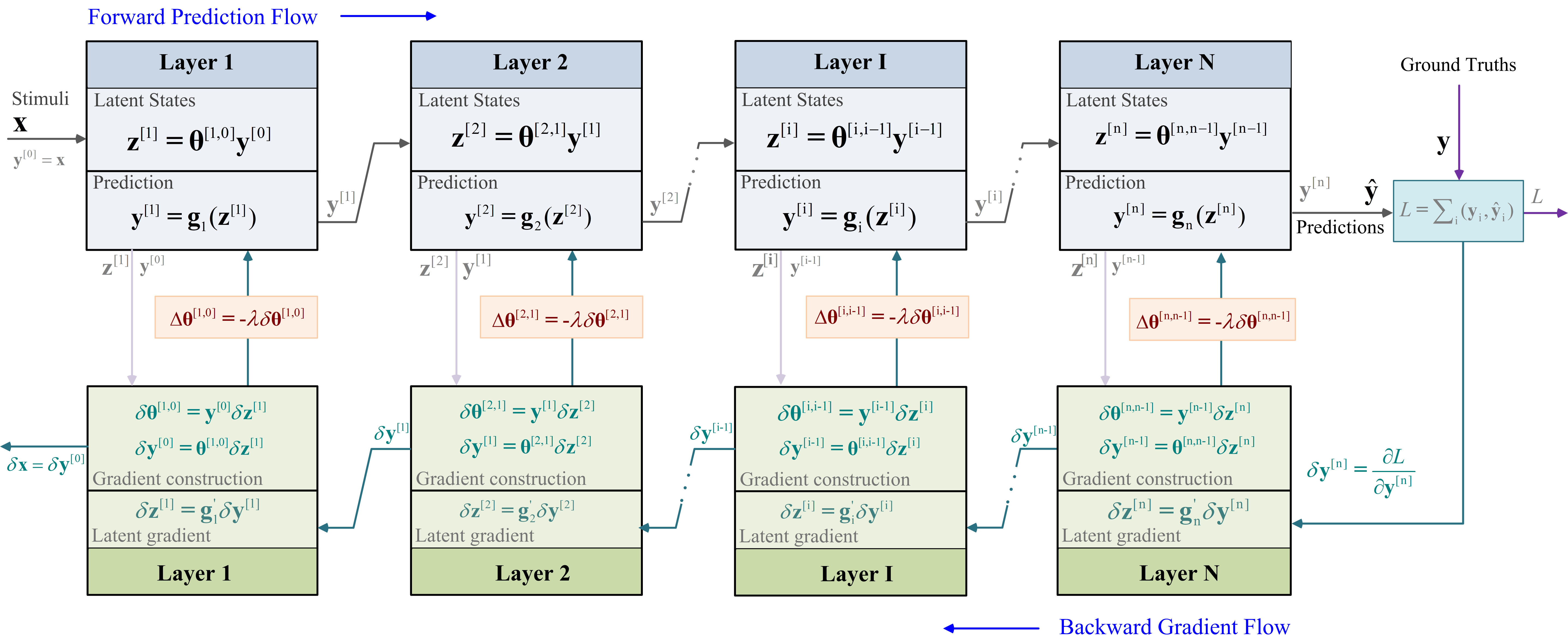}\\[0.8ex]
			b) Back-Propagation Structured Offline Training Gradient Flow
		\end{tabular}
	\end{center}
	\vspace{0.3em}
	\caption{On-line Inference and Off-line Training Framework of Deep Learning Models}
	\label{fig:layered-global}
\end{figure}

Fig.~\ref{fig:layered-global} expands this global-learning topology. During online inference, the input is transformed sequentially through the layer hierarchy, and each intermediate representation is supplied only to the next forward layer. During offline training, a loss at the terminal output initiates a reverse gradient path through the same hierarchy. The update of an early parameter block therefore depends on sensitivities accumulated across all downstream layers. The architecture scales representational capacity effectively, but it also separates online operation from the phase in which a globally supplied objective can modify the model.

Under this supervised-learning paradigm, a deep model extracts statistical knowledge from data when the samples are correctly labeled and the network has sufficient capacity. The limitation appears when the deployed environment is not represented in the training data. The separation between offline training and online inference prevents the system from modifying its model in direct response to new physical evidence. Deep networks nevertheless show that systems with billions of parameters can be optimized in a stable and repeatable manner. Their hierarchical structure makes systematic application of backpropagation possible. If the same structured capacity could be combined with stable online learning, the adaptability of these systems would change fundamentally.

\begin{figure}[!htb]
	\centering
	\includegraphics[scale=.5]{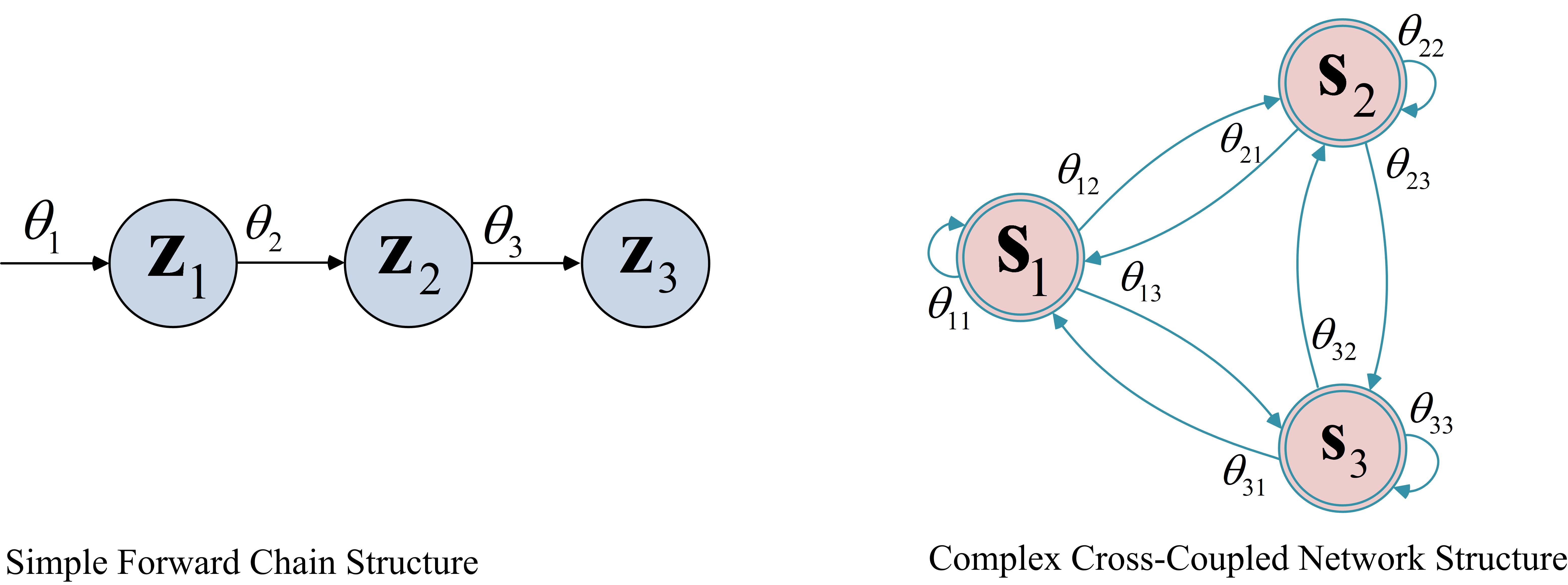}
	\caption{Model Structure Comparison for Deep learning and Adaptive learning}\label{fig:structure}
\end{figure}

Fig.~\ref{fig:structure} emphasizes the structural contrast. A deep network is represented as a simple forward chain, whereas a conventional adaptive model contains cross-referenced states and internal feedback.

Adaptive learning, shown in Fig.~\ref{fig:paradigms}(b), (d), and (f), follows the principles of adaptive control \cite{aastrom1995adaptive,ioannou1996robust}. Instead of a static algebraic map, it uses a dynamic state-space model,

\begin{equation}
\begin{array}{rcl}
\dot{\hat{\mathbf{s}}} & = & f(\hat{\mathbf{s}},\bm{\uptheta},\mathbf{x}) \\
\hat{\mathbf{y}} & = & g(\hat{\mathbf{s}},\bm{\uptheta})
\end{array}
\end{equation}

where \(\hat{\mathbf{s}}=[\hat{s}_1,\hat{s}_2,\ldots,\hat{s}_n]^T\) contains the estimated states, \(\bm{\uptheta}=[\uptheta_1,\uptheta_2,\ldots,\uptheta_k]^T\) contains the parameters, and \(\hat{\mathbf{y}}=[\hat{y}_1,\hat{y}_2,\ldots,\hat{y}_m]^T\) contains the predicted outputs. Typically \(n\gg m\) and \(n\gg k\). The model is therefore more complex than the compact state equations suggest. State interactions form internal nonlinear feedback loops and cross-references with widely varying frequency characteristics.

The intrinsic principle of adaptive learning is continuous feedback of the output prediction error,

\begin{equation}
\textbf{e}_y=\textbf{y}_s-\hat{\textbf{y}}_s
\end{equation}

where \(\textbf{y}_s\) and \(\hat{\textbf{y}}_s\) are the measured and predicted outputs. Parameters are adjusted according to

\begin{equation*}
\dot{\bm{\uptheta}}=\lambda h(\textbf{e}_y,\hat{\mathbf{s}}).
\end{equation*}

Ideally, the dynamic model approaches the physical system and \(\bm{\uptheta}\) approaches the physical parameter \(\bm{\uptheta}_{\mathrm{GT}}\) as \(\textbf{e}_y\rightarrow0\). For a large nonlinear system, however, the solution to \(\textbf{e}_y=0\) may not be unique, and an arbitrary adaptation law does not ensure global stability. If the mapping from \(\uptheta\cdot\hat{\mathbf{s}}\) to \(\textbf{e}_y\) is strictly positive real, then the law

\begin{equation}
\dot{\bm{\uptheta}}=\lambda\textbf{e}_y\hat{\mathbf{s}}
\end{equation}

can ensure learning stability and solution uniqueness \cite{slotine1991applied}. This condition is often violated in complex adaptive systems. The update law is also intrinsically related to gradient-based optimization.

Adaptive learning has the important advantage of unsupervised online operation. It can adapt to a changing world without a manually labeled data set or a complete offline retraining stage. The distinction is conceptual as well as computational. Deep learning primarily extracts statistical structure from data, whereas adaptive learning attempts to identify the physical process directly from sensor measurements in real time. Both nevertheless depend on feedback: deep learning propagates a predefined loss, and adaptive learning propagates a tracking error.

The main constraint of adaptive learning is scalability. Unlike hierarchical deep networks, conventional adaptive systems contain parameters that are heavily coupled by internal cross-references. As the parameter space grows, global convergence becomes mathematically intractable, and strict proofs are normally possible for only a limited number of variables \cite{aastrom1995adaptive,ioannou1996robust,slotine1991applied}. This creates a trade-off between model complexity and analytical rigor. The purpose of OAL is to remove that trade-off by combining the structured hierarchy of deep learning with the real-time error-feedback principle of adaptive learning.

\FloatBarrier
\subsection{Conceptual Online Active Learning Framework}\label{conceptual-online-active-learning-framework}

Adaptive learning can identify physical dynamics in real time, but it does not scale easily because its representation of the physical world is unstructured. Deep learning can train very large neural systems, but its online use is restricted by the dependence on labeled objectives and separated training. OAL synthesizes these paradigms to obtain continuous learning in a large structured network.

\begin{figure}[h]
\centering
\includegraphics[scale=.43]{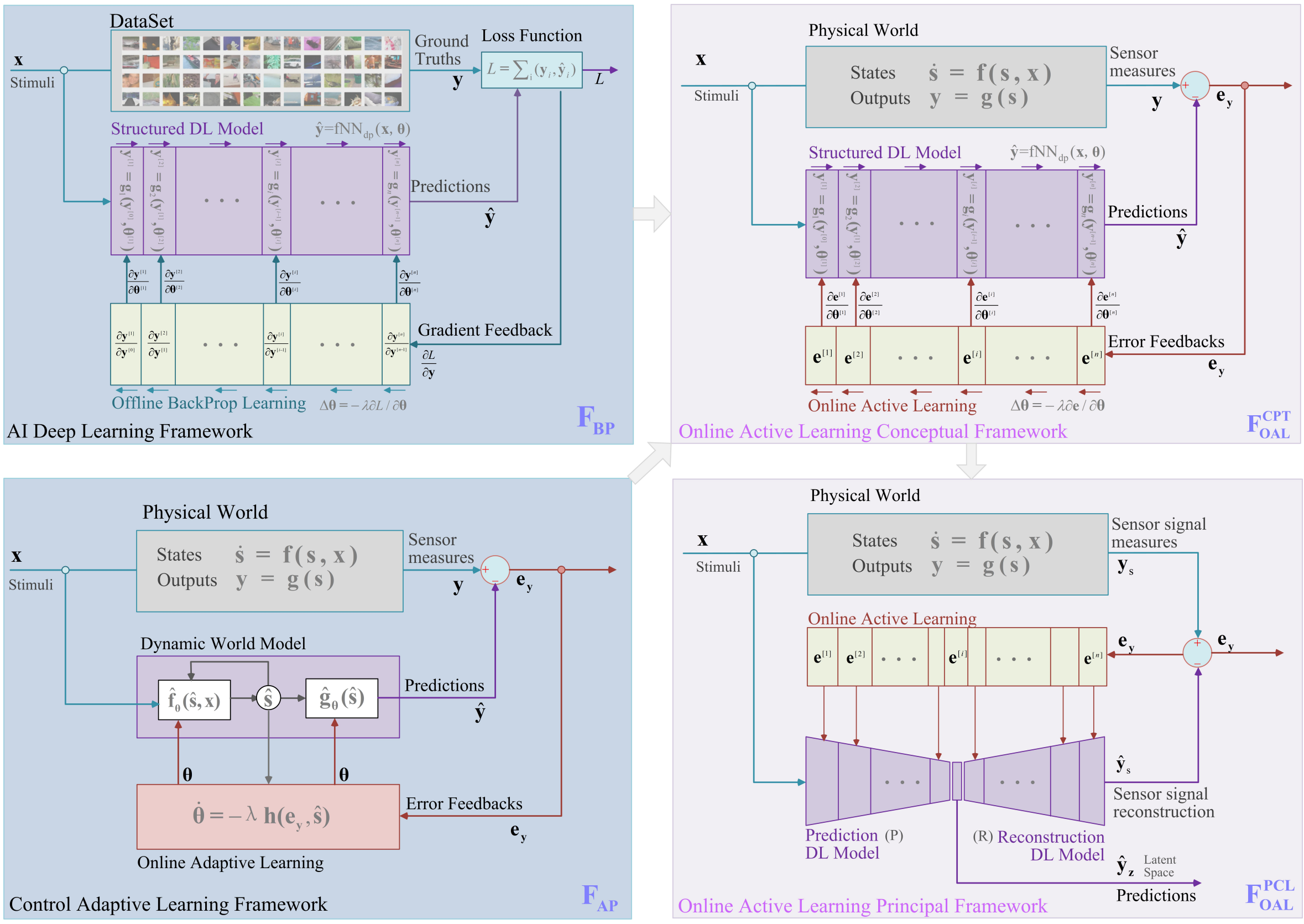}
\caption{Conceptual Active Learning Framework}\label{fig:conceptual}
\end{figure}

The scalability problem of adaptive learning originates in the nonlinear dynamics of Eq.~(5). Directly approximating the entire state evolution with a deep network is not sufficient, even though residual networks resemble discretized ordinary differential equations \cite{he2016deep}. A physical process is nevertheless dissipative and therefore admits a stable realized solution. If that solution is written as

\begin{equation}
\hat{\mathbf{s}}=f_\textrm{sln}(\bm{\uptheta},\mathbf{x})
\end{equation}

the hierarchical representation in Eq.~(1) can approximate the solved input-state-output relationship. The representation can first be organized by conventional gradient learning and can then be embedded in a live feedback loop. Replacing the classical physics-based adaptive model with this structured data-driven neural system produces the conceptual OAL framework in Fig.~\ref{fig:conceptual}.

\FloatBarrier

The framework contains four components.

\textbf{Physical world.} The environment is a dynamic system in which stimulus \(\mathbf{x}\) acts on state \(\mathbf{s}\):

\begin{equation*}
\dot{\mathbf{s}}=f(\mathbf{s},\mathbf{x}),\qquad\mathbf{y}=g(\mathbf{s}).
\end{equation*}

The sensors produce the measurement \(\mathbf{y}\), which acts as the current empirical reference against which the predictive model is evaluated.

\textbf{Structured deep-learning model.} A hierarchical neural network represents the external physical process:

\begin{equation*}
\hat{\mathbf{y}}=f{\mathbb{NN}}_\mathrm{dl}(\mathbf{x},\bm{\uptheta}).
\end{equation*}

Its layers \(\mathbf{y}^{[1]},\mathbf{y}^{[2]},\ldots,\mathbf{y}^{[n]}\) provide sufficient capacity to approximate a nonlinear environment. The same layer organization also makes it possible to apply parameter updates at specific depths.

\textbf{Error measurement.} The evaluator computes the instantaneous discrepancy between the measurement and prediction:

\begin{equation*}
\textbf{e}_y=\mathbf{y}_s-\hat{\mathbf{y}}_s.
\end{equation*}

The feedback loop minimizes the prediction-error energy

\begin{equation}
\mathcal{L}_e=\frac{1}{2}\textstyle\sum\nolimits_k(\mathbf{y}_{k,s}-\hat{\mathbf{y}}_{k,s})^2
=\frac{1}{2}\textstyle\sum\nolimits_k e_k^2
=\frac{1}{2}\textbf{e}_y^T\textbf{e}_y
\end{equation}

Continuous reduction of this energy is consistent with the minimum-energy interpretation used in neuroscience \cite{friston2009theory}.

\textbf{Online active learning.} The terminal residual is transmitted through the hierarchy to obtain layer-specific errors \(\textbf{e}^{[1]},\textbf{e}^{[2]},\ldots,\textbf{e}^{[n]}\):

\begin{equation}
\textbf{e}^{[i]}=\textbf{J}_y^{[i,i+1]}\textbf{e}^{[i+1]}
\end{equation}

where

\begin{equation}
\textbf{J}_y^{[i,i+1]}=\frac{\partial\mathbf{y}^{[i+1]}}{\partial\mathbf{y}^{[i]}}
\end{equation}

The notation \(\textbf{J}_y^{[i]}\) is used for the same layer sensitivity where the two indices are unnecessary. OAL then applies the local parameter law

\begin{equation}
\Delta\bm{\uptheta}^{[i]}=\lambda_i\textbf{J}_\uptheta^{[i]}\mathbf{e}^{[i]}
\end{equation}

in which \(\lambda_i\) is the learning rate and

\begin{equation}
\textbf{J}_\uptheta^{[i]}=\partial{\mathbf{y}^{[i]}}/\partial\bm{\uptheta}^{[i]}
\end{equation}

is the sensitivity of the layer output to its parameters. The Jacobian matrix specifies the localized sensitivity function. To express the complete backward path from the terminal output to layer \(i\), define

\begin{equation}
\mathbb{J}_y^{[i]}=\prod_{k=i}^{n}\textbf{J}_y^{[k]}
\end{equation}

Because \(\textbf{J}_y^{[n]}=\mathbf{I}\), the product to layer \(n\) is equivalent to the product to \(n-1\). The corresponding sensitivity from the terminal output to \(\bm{\uptheta}^{[i]}\) is

\begin{equation}
\mathbb{J}_\uptheta^{[i]}=\textbf{J}_\uptheta^{[i]}\mathbb{J}_y^{[i]}
\end{equation}

The global form of the learning law is therefore

\begin{equation}
\Delta\bm{\uptheta}^{[i]}=\lambda_i\mathbb{J}_\uptheta^{[i]}\mathbf{e}_y
\end{equation}

The OAL paradigm uses the prediction error to correct the neural approximation continuously. The capacity of the DNN allows it to represent a complex nonlinear external process, while real-time learning becomes iterative optimization of \(\bm{\uptheta}\) driven by physical residuals rather than by newly supplied labels.

Two observations clarify the stability direction of this conceptual loop. First, \(L_e\) is a Lyapunov-like energy function. Its derivative is

\begin{equation}
\begin{array}{rcl}
\dot{\mathcal{L}}_e &=& \textbf{e}_y^T\dot{\textbf{e}}_y\\
&=&\textbf{e}_y^T\left(\dot{\textbf{y}}_s-\dot{\hat{\textbf{y}}}_s\right)\\
&=&\textbf{e}_y^T\left(\dot{\textbf{y}}_s-\frac{\partial f_\mathrm{NN}^{\;\;\;T}}{\partial\mathbf{x}}\dot{\mathbf{x}}-\frac{\partial f_\mathrm{NN}^{\;\;\;T}}{\partial\bm{\uptheta}}\dot{\bm{\uptheta}}\right)\\
&=&\textbf{e}_y^T\left(\dot{\textbf{y}}_s-\frac{\partial f_\mathrm{NN}^{\;\;\;T}}{\partial\mathbf{x}}\dot{\mathbf{x}}-\lambda\frac{\partial f_\mathrm{NN}^{\;\;\;T}}{\partial\bm{\uptheta}}\frac{\partial f_\mathrm{NN}}{\partial\bm{\uptheta}}\textbf{e}_y\right)
\end{array}
\end{equation}

When \(\dot{x}\) and \(\dot{y}_s\) are treated as bounded external terms and their difference is negligible relative to the learning dynamics, the expression becomes

\begin{equation}
\dot{L}_e=-\lambda\textbf{e}_y^T\left(\frac{\partial f_\mathrm{NN}}{\partial\bm{\uptheta}}\right)^{T}
\frac{\partial f_\mathrm{NN}}{\partial\bm{\uptheta}}\textbf{e}_y\leq0
\end{equation}

The negative-semidefinite energy derivative provides the local convergence direction. Second, the positive-semidefinite term

\begin{equation*}
\textbf{J}_{\uptheta}^T\textbf{J}_{\uptheta}\geq0
\end{equation*}

does not require the exact update of Eq.~(12). A computationally simpler law can use only the sign of the sensitivity:

\begin{equation}
\Delta\bm{\uptheta}^{[i]}=\lambda_i\textrm{sgn}(\mathbb{J}_\uptheta^{[i]})\mathbf{e}^{[i]}
\end{equation}

The sign-based law avoids requiring complete analytical knowledge of the Jacobian at every iteration. For scalar or component-wise decoupled sensitivities, the sign operation preserves the descent direction. Its extension to coupled matrix-valued Jacobians requires additional structural conditions. These relations state the architectural stability direction necessary for constructing the feedback loop. The complete high-dimensional proof, including the matrix-valued conditions and the conditions under which many local loops compose into a stable global learning system, is developed in the companion stability paper.

\FloatBarrier
\subsection{Principle of Online Active Learning}\label{principle-of-online-active-learning}

The conceptual framework still has two limitations. First, minimizing only the discrepancy between \(\hat{\textbf{y}}_s\) and \(\textbf{y}_s\) can correct the observable output without producing a meaningful latent representation. Second, asymptotic output tracking does not by itself guarantee that all model parameters remain well behaved. Parameter stability is more fundamental than output convergence from the viewpoint of system dynamics. The present section addresses representation learning and states the local update structure; the full parameter-stability analysis is delegated to the companion paper.

\begin{figure}[h]
\centering
\includegraphics[scale=.43]{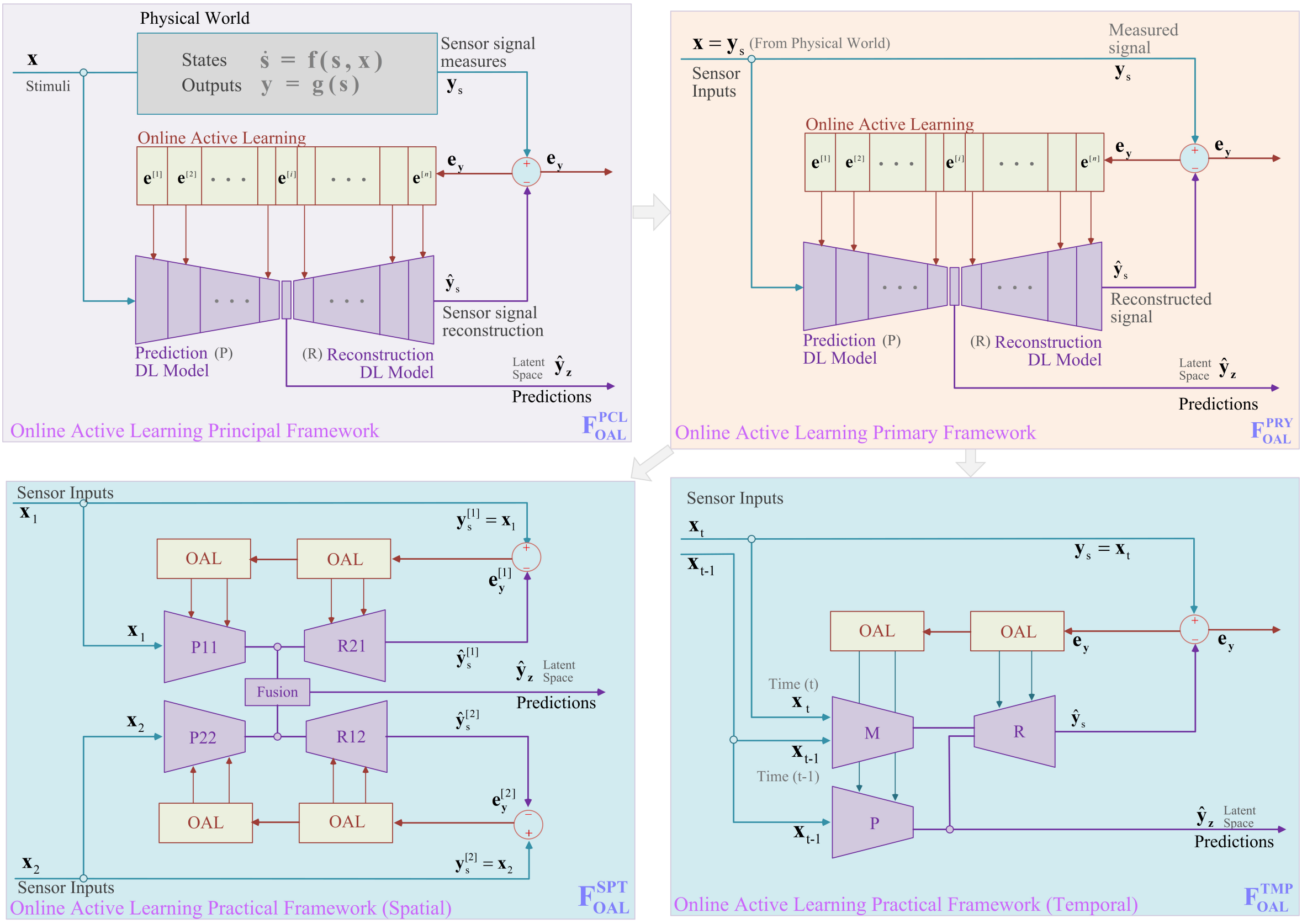}
\caption{Primary Active Learning Frameworks}\label{fig:primary}
\end{figure}

Fig.~\ref{fig:primary} summarizes the progression from the principal architecture to the primary, spatial, and temporal frameworks.

\FloatBarrier
\subsubsection{Principal Prediction Latent and Reconstruction Architecture}\label{principal-prediction-latent-and-reconstruction-architecture}

To obtain a meaningful representation, the principal framework decomposes the monolithic DNN into functionally distinct prediction, latent, and reconstruction models.

\textbf{Prediction model P.} The forward network maps high-dimensional raw input \(\mathbf{x}\in\mathbb{R}^{D}\) to a lower-dimensional latent manifold \(\hat{\mathbf{y}}_z\in\mathbb{R}^{d}\), where \(d\ll D\):

\begin{equation}
\hat{\mathbf{y}}_z=f{\mathbb{NN}}_\mathrm{dP}(\mathbf{x},\bm{\uptheta}_P)
\end{equation}

The compressive map is intended to isolate low-rank structural invariants in the data.

\textbf{Latent model L.} The latent network manages information flow and state transitions inside the compact representation:

\begin{equation}
\hat{\mathbf{y}}_l=f{\mathbb{NN}}_\mathrm{dL}(\hat{\mathbf{y}}_z,\bm{\uptheta}_L)
\end{equation}

Its parameter set \(\bm{\uptheta}_L\) maps internal dynamics without returning immediately to the high-dimensional observation space.

\textbf{Reconstruction model R.} The generative network maps the evolved latent representation back to the observation manifold:

\begin{equation}
\hat{\mathbf{y}}_s=f{\mathbb{NN}}_\mathrm{dR}(\hat{\mathbf{y}}_l,\bm{\uptheta}_R)
\end{equation}

The reconstructed signal has the same dimension as the physical measurement and can therefore be evaluated directly. For structural simplicity, \(L\) may be incorporated into \(P\), incorporated into \(R\), or distributed across them. Symmetric inverted topologies for \(P\) and \(R\) also enable cross-layer routing and skip connections. Such symmetry shortens error paths and improves the efficiency of representation learning during online updates.

\FloatBarrier
\subsubsection{Progression from Principal to Practical Frameworks}\label{progression-from-principal-to-practical-frameworks}

The architectural progression in Fig.~\ref{fig:primary} moves from idealized state tracking toward deployable representation learning. It addresses the tension between accurate signal reconstruction and expressive latent variables by changing topology and information routing.

The \textbf{principal framework} assumes some knowledge of the external state \(\mathbf{s}\) and its transition \(\dot{\mathbf{s}}=f(\mathbf{s},\mathbf{x})\). Those quantities are rarely available directly in an operating system. The \textbf{primary framework} moves the system boundary to the sensor stream by setting \(\mathbf{x}=\mathbf{y}_s\). The resulting pipeline is data driven and can be deployed at the sensory boundary. This arrangement resembles a Markov blanket: internal representation and model correction are driven by signals that cross the sensing interface \cite{kirchhoff2018markov}.

The closed loop of the primary framework introduces a mathematical vulnerability. Because the input and reconstruction target are identical, the system may converge to the trivial inverse solution \(\mathbb{R}=\mathbb{P}^{-1}\). That solution can yield zero output error without representing physical structure. The \textbf{spatial} and \textbf{temporal} frameworks therefore break the symmetry in different ways. The spatial framework uses cross-channel reconstruction; the temporal framework predicts across a time interval. Both constrain the network to model relationships that cannot be solved by copying the input.

\FloatBarrier
\subsubsection{Primary Framework and Variational Energy}\label{primary-framework-and-variational-energy}

The primary framework sets the system input equal to the sensor stream, \(\mathbf{x}=\mathbf{y}_s\), so that information about the environment is obtained through the sensing boundary. The prediction residual becomes

\begin{equation*}
\textbf{e}_s=\mathbf{y}_s-\hat{\mathbf{y}}_s,
\end{equation*}

and the prediction model is

\begin{equation}
\hat{\mathbf{y}}_z=f{\mathbb{NN}}_\mathrm{dP}(\mathbf{y}_s,\bm{\uptheta}_P)
\end{equation}

Assuming Gaussian uncertainty in the environmental channel, the complete parameter vector \(\bm{\uptheta}=[\bm{\uptheta}_P,\bm{\uptheta}_L,\bm{\uptheta}_R]^T\) is updated by minimizing the regularized variational energy

\begin{equation}
\mathcal{L}_e=\frac{1}{2}\textbf{e}_s^T\mathbf{W}_y\textbf{e}_s+
\frac{1}{2}\bm{\uptheta}^T\mathbf{W}_\uptheta\bm{\uptheta}
\end{equation}

where \(\mathbf{W}_y\) represents sensory precision and \(\mathbf{W}_{\uptheta}\) represents parameter regularization. Continuous online adaptation of the reconstruction model follows

\begin{equation}
\dot{\bm{\uptheta}}_R+\kappa_R\bm{\uptheta}_R
=\lambda_R\textbf{J}_{\uptheta_R}^T\textbf{e}_R
\end{equation}

The effective rate \(\lambda_R\) combines the learning coefficient with sensory precision. The learning-inertia coefficient \(\kappa_R\) is determined by parameter regularization and the internal learning dynamics. It preserves useful internal priors while providing damping that improves real-time stability. The Jacobian \(\textbf{J}_{\uptheta_R}\) guides the top-down residual into the reconstruction parameter space.

Unrolling the error through the three models gives the latent-model update

\begin{equation}
\textbf{e}_L=\textbf{J}_{L}\textbf{e}_R;\qquad
\dot{\bm{\uptheta}}_L+\kappa_L\bm{\uptheta}_L
=\lambda_L\textbf{J}_{\uptheta_L}\textbf{e}_L
\end{equation}

and the prediction-model update

\begin{equation}
\textbf{e}_P=\textbf{J}_{P}\textbf{e}_L;\qquad
\dot{\bm{\uptheta}}_P+\kappa_P\bm{\uptheta}_P
=\lambda_P\textbf{J}_{\uptheta_P}\textbf{e}_P
\end{equation}

The closed error path converts a label-dependent offline network into an autonomous self-correcting system. The residual \(\textbf{e}_s\) continuously reshapes the internal predictive models in response to environmental changes and sensor anomalies.

\FloatBarrier
\subsubsection{Spatial Framework}\label{spatial-framework}

The primary framework can fail through the identity solution. If \(\mathbb{R}=\mathbb{P}^{-1}\), then \(\hat{\mathbf{y}}_s=\mathbf{y}_s\) and \(\mathbf{e}_y=\mathbf{0}\), but the system may have extracted no physical knowledge. The spatial OAL framework prevents this degeneration by cross-routing two different sensory channels, \(\mathbf{y}_s^{[1]}\) and \(\mathbf{y}_s^{[2]}\). Two prediction models map the channels into separate latent spaces:

\begin{equation}
\hat{\mathbf{y}}_z^{[1]}=f{\mathbb{NN}}_\mathrm{dP}(\mathbf{y}_s^{[1]},\bm{\uptheta}_{P_{11}});\quad
\hat{\mathbf{y}}_z^{[2]}=f{\mathbb{NN}}_\mathrm{dP}(\mathbf{y}_s^{[2]},\bm{\uptheta}_{P_{22}})
\end{equation}

The reconstruction models are cross-coupled, so each branch predicts the alternative observation:

\begin{equation}
\hat{\mathbf{y}}_s^{[1]}=f{\mathbb{NN}}_\mathrm{dR}(\hat{\mathbf{y}}_z^{[2]},\bm{\uptheta}_{R_{21}});\quad
\hat{\mathbf{y}}_s^{[2]}=f{\mathbb{NN}}_\mathrm{dR}(\hat{\mathbf{y}}_z^{[1]},\bm{\uptheta}_{R_{12}})
\end{equation}

The two tracking errors are

\begin{equation*}
\mathbf{e}_y^{[1]}=\mathbf{y}_s^{[1]}-\hat{\mathbf{y}}_s^{[1]};\quad
\mathbf{e}_y^{[2]}=\mathbf{y}_s^{[2]}-\hat{\mathbf{y}}_s^{[2]}.
\end{equation*}

The pair \([\bm{\uptheta}_{R_{12}},\bm{\uptheta}_{P_{11}}]\) adapts according to \(\mathbf{e}_y^{[2]}\), while \([\bm{\uptheta}_{R_{21}},\bm{\uptheta}_{P_{22}}]\) adapts according to \(\mathbf{e}_y^{[1]}\). Because \(\mathbb{R}_{21}\) must reconstruct Sensor 1 using a latent representation derived from Sensor 2, neither branch can collapse into a local identity map. The latent features may be concatenated and fused by a dedicated network,

\begin{equation}
\hat{\mathbf{y}}_{z}=\mathcal{G}_{\text{fuse}}(\hat{\mathbf{y}}_{z}^{[1]},\hat{\mathbf{y}}_{z}^{[2]})
\end{equation}

so that both channels contribute to the task representation. Cross-channel prediction forces the system to internalize invariant spatial correlations and physical boundaries between coordinate spaces. The design is related to binocular processing, in which two non-identical observations are fused into a coherent environmental representation \cite{tonioni2019learning}.

\FloatBarrier
\subsubsection{Temporal Framework}\label{temporal-framework}

The temporal OAL framework breaks the identity solution by displacing the optimization target across time. It uses a prediction model \(\mathbb{P}\), a motion model \(\mathbb{M}\), and a reconstruction model \(\mathbb{R}\). The prediction model extracts the latent representation of the historical measurement:

\begin{equation}
\hat{\mathbf{y}}_{z,t-1}=f{\mathbb{NN}}_\mathrm{dP}(\mathbf{y}_{s,t-1},\bm{\uptheta}_P)
\end{equation}

The same prediction model provides the current latent representation:

\begin{equation*}
\hat{\mathbf{y}}_{z,t}=f{\mathbb{NN}}_\mathrm{dP}(\mathbf{y}_{s,t},\bm{\uptheta}_P).
\end{equation*}

The motion model represents the state transition over \([t-1,t]\) and infers relative motion in latent space:

\begin{equation}
\mathbf{m}_{t,t-1}=\mathbf{F}_t-\mathbf{F}_{t-1}
=f{\mathbb{NN}}_\mathrm{dM}(\hat{\mathbf{y}}_{z,t},\hat{\mathbf{y}}_{z,t-1},\bm{\uptheta}_M)
\end{equation}

where \(\mathbf{F}_t\) and \(\mathbf{F}_{t-1}\) are structural features at the two instants. The reconstruction model combines the historical latent state and inferred motion to predict the current observation:

\begin{equation}
\hat{\mathbf{y}}_{s,t}=f{\mathbb{NN}}_\mathrm{dR}(\hat{\mathbf{y}}_{z,t-1},\mathbf{m}_{t,t-1},\bm{\uptheta}_R)
\end{equation}

The parameter set \([\bm{\uptheta}_P,\bm{\uptheta}_M,\bm{\uptheta}_R]\) is optimized from

\begin{equation*}
\mathbf{e}_{s,t}=\mathbf{y}_{s,t}-\hat{\mathbf{y}}_{s,t}.
\end{equation*}

Because the current observation is reconstructed from the historical latent state and the relative motion estimated from the current and historical representations, the system cannot solve the task through a static inverse \(\mathbb{R}=\mathbb{P}^{-1}\). The temporal displacement requires the latent model to capture properties such as causal order, change, velocity, and inertia.

\FloatBarrier
\subsection{Practical Extensions of the Online Active Learning Framework}\label{practical-extensions-of-the-online-active-learning-framework}

\begin{figure}[!h]
	\centering
	\includegraphics[scale=.44]{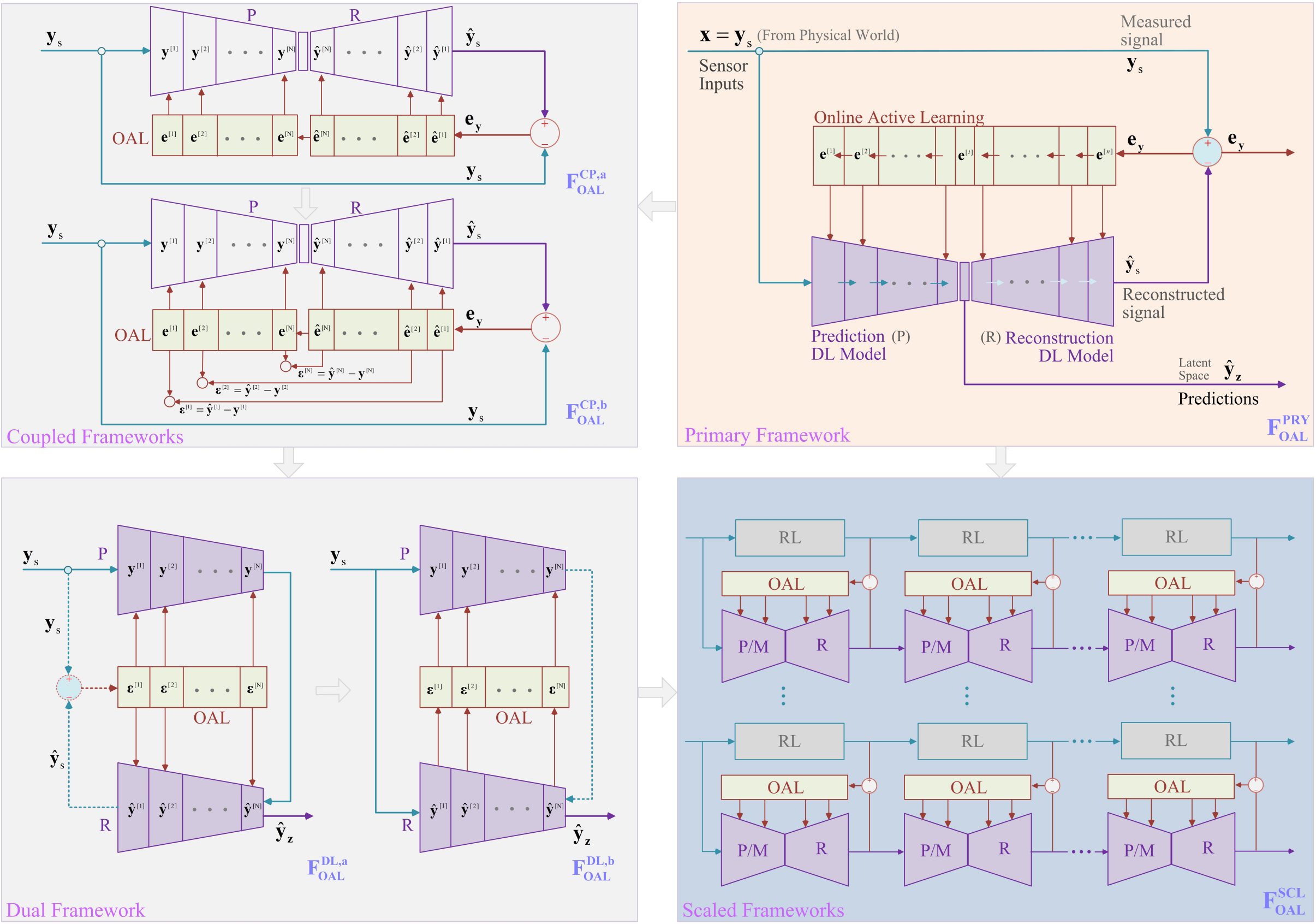}
	\caption{Practical Active Learning Frameworks}\label{fig:practical}
\end{figure}

The primary OAL architecture in Figs.~\ref{fig:primary} and \ref{fig:practical} performs end-to-end online learning. All parameter adaptations originate from the global error \(\textbf{e}_y\) at the terminal layer between the measured and reconstructed signals. Although the feedback direction is established above and the complete stability proof is given separately, the end-to-end architecture imposes constraints that differ from conventional offline learning, including active damping and appropriate activation properties. More importantly, its learning efficiency is limited by the long pathway for backward error transmission. When the terminal error reaches an intermediate layer, products of many Jacobians may cause severe vanishing or exploding effects.

Localized learning avoids the extensive backward path by generating errors \(\textbf{e}_i\) close to the parameter blocks that they update. This both supports online operational stability and accelerates learning. The practical architectures in Fig.~\ref{fig:practical} therefore preserve the prediction-reconstruction principle while progressively localizing the error mechanics.


\FloatBarrier

\subsubsection{Coupled Dual Framework}\label{coupled-dual-framework}

The coupled framework first reorganizes \(\mathbb{P}\) and \(\mathbb{R}\) into a reciprocal dual system. The prediction model maps inputs forward through

\begin{equation}
\mathbf{y}^{[i]}=f_{\mathbb{NN}}^{[i]}(\mathbf{y}^{[i-1]},\bm{\uptheta}^{[i]}),\qquad
\mathbf{y}^{[1]}=f_{\mathbb{NN}}^{[1]}(\mathbf{y}_s,\bm{\uptheta}^{[1]})
\end{equation}

for \(i=1,2,\ldots,N\). The variables \(\mathbf{y}^{[1]},\mathbf{y}^{[2]},\ldots,\mathbf{y}^{[N]}\) are the latent representations and \(\bm{\uptheta}^{[1]},\bm{\uptheta}^{[2]},\ldots,\bm{\uptheta}^{[N]}\) are the corresponding parameters. The reconstruction model contains functionally matching layers arranged in reverse topological order:

\begin{equation}
\hat{\mathbf{y}}^{[i]}=f_{\mathbb{NN}}^{[i]}(\hat{\mathbf{y}}^{[i+1]},\bm{\Phi}^{[i]}),\qquad
\hat{\mathbf{y}}^{[N]}=f_{\mathbb{NN}}^{[N]}(\mathbf{y}^{[N]},\bm{\Phi}^{[N]})
\end{equation}

for \(i=N,N-1,\ldots,1\). It produces the latent reconstructions \(\hat{\mathbf{y}}^{[N]},\hat{\mathbf{y}}^{[N-1]},\ldots,\hat{\mathbf{y}}^{[1]}\) under parameters \(\bm{\Phi}^{[N]},\bm{\Phi}^{[N-1]},\ldots,\bm{\Phi}^{[1]}\). The reconstructed sensor prediction is \(\hat{\mathbf{y}}_s=\hat{\mathbf{y}}^{[1]}\).

\Needspace{12\baselineskip}
The mirror topology yields reciprocal error transmission. The backward signal through \(R\) is

\begin{equation}
\hat{\mathbf{e}}^{[i]}=\mathbf{J}_{\hat{y}}^{[i]}\hat{\mathbf{e}}^{[i-1]},\qquad
\hat{\mathbf{e}}^{[1]}=\mathbf{e}_y
\end{equation}

for \(i=2,\ldots,N\), with reconstruction update

\begin{equation*}
\Delta\bm{\Phi}^{[i]}=\lambda_i\mathbf{J}_{\Phi}^{[i]}\hat{\mathbf{e}}^{[i]}.
\end{equation*}

Error propagation through \(P\) is

\begin{equation}
\mathbf{e}^{[i]}=\mathbf{J}_y^{[i]}\mathbf{e}^{[i+1]},\qquad
\mathbf{e}^{[N]}=\mathbf{J}_{y\hat{y}}\hat{\mathbf{e}}^{[N]}
\end{equation}

for \(i=N-1,N-2,\ldots,1\), with prediction update

\begin{equation*}
\Delta\bm{\uptheta}^{[i]}=\lambda_i\mathbf{J}_{\uptheta}^{[i]}\mathbf{e}^{[i]}.
\end{equation*}

To introduce localized error mechanics, define an intrinsic error at each paired layer:

\begin{equation}
\boldsymbol{\varepsilon}^{[i]}=\hat{\mathbf{y}}^{[i]}-\mathbf{y}^{[i]},\qquad i=1,2,\ldots,N
\end{equation}

The total prediction error combines the global and local components:

\begin{equation}
\mathbf{e}_{\delta}^{[i]}=\eta\mathbf{e}^{[i]}+(1-\eta)\boldsymbol{\varepsilon}^{[i]},\qquad0\leq\eta\leq1
\end{equation}

and the prediction parameter update becomes

\begin{equation}
\Delta\bm{\uptheta}^{[i]}=\lambda_i\mathbf{J}_{\uptheta}^{[i]}\mathbf{e}_{\delta}^{[i]}
\end{equation}

The local component lets \(P\) use the immediate representation supplied by the corresponding layer of \(R\), shortening the credit-assignment path and improving learning efficiency.

\FloatBarrier
\subsubsection{Decoupled Dual Framework}\label{decoupled-dual-framework}

With only terminal feedback, the contribution of every parameter layer to the total error depends on the corresponding long-chain Jacobian. Here \(\delta\mathbf{e}_y^{[k]}\) denotes the one-step variation in the terminal prediction error caused by the parameter update at iteration \(k\). Partitioning this variation across the individual layers under their corresponding learning laws yields

\begin{equation}
\delta\mathbf{e}_y^{[k]}=-\left[
\lambda_1\mathbb{J}_{\uptheta}^{[1]T}\mathbb{J}_{\uptheta}^{[1]}
+\lambda_2\mathbb{J}_{\uptheta}^{[2]T}\mathbb{J}_{\uptheta}^{[2]}
+\cdots+\lambda_i\mathbb{J}_{\uptheta}^{[i]T}\mathbb{J}_{\uptheta}^{[i]}
+\cdots+\lambda_n\mathbb{J}_{\uptheta}^{[n]T}\mathbb{J}_{\uptheta}^{[n]}
\right]\mathbf{e}_y^{[k]}
\end{equation}

The magnitude of \(\mathbb{J}_{\uptheta}^{[i]}\), which includes the chained sensitivity from layer \(i\) to the output, may make a layer extremely sensitive or almost insensitive to \(\textbf{e}_y\). Terminal error is therefore inefficient for regulating every internal parameter. Localized learning instead uses

\begin{equation}
\Delta\bm{\uptheta}^{[i]}=\lambda_i\mathbf{J}_{\uptheta}^{[i]}\boldsymbol{\varepsilon}^{[i]}
\end{equation}

The resulting local error variation is

\begin{equation}
\delta\boldsymbol{\varepsilon}_{[k+1]}^{[i]}=
-\lambda_i\mathbf{J}_{\uptheta}^{[i]T}
\mathbf{J}_{\uptheta}^{[i]}\boldsymbol{\varepsilon}_{[k]}^{[i]}
\end{equation}

Here \(\delta\boldsymbol{\varepsilon}_{[k+1]}^{[i]}\) denotes the corresponding one-step variation in the local residual of layer \(i\).

The local residual offers more direct controllability and supports parallel layer-wise optimization. The monolithic coupled structure consequently advances to the decentralized dual architecture summarized in Fig.~\ref{fig:practical}.

The prediction and reconstruction branches preserve the dual relationship

\begin{equation*}
\mathbf{y}^{[i]}=f_{\mathbb{NN}}^{[i]}(\mathbf{y}^{[i-1]},\bm{\uptheta}^{[i]}),\qquad
\hat{\mathbf{y}}^{[i]}=f_{\mathbb{NN}}^{[i]}(\hat{\mathbf{y}}^{[i+1]},\bm{\Phi}^{[i]}).
\end{equation*}

When the forward and backward parameters satisfy an appropriate transposition condition,

\begin{equation}
\bm{\Phi}^{[i]}=\left(\bm{\uptheta}^{[i]}\right)^T
\end{equation}

online optimization can be concentrated in one branch while the other supplies the dual reconstruction. Two boundary variants are possible. In the recommended form, \(\mathbb{R}\) is driven backward by the terminal latent state \(\mathbf{y}^{[N]}\), so that the first-layer local residual coincides with the sensor error. In the alternative form, \(\mathbb{R}\) is driven forward by the source signal \(\mathbf{y}_s\), which creates a duplicated inference path and is therefore not recommended.

The recommended dual architecture resembles biological predictive coding, but it differs in an important respect. Standard predictive-coding schemes require additive learning of the latent representation \(\hat{\mathbf{y}}^{[i]}\). The present dual architecture removes that additional latent-state update and maintains a layer-wise isomorphism between each latent state \(\mathbf{y}^{[i]}\) and reconstruction \(\hat{\mathbf{y}}^{[i]}\).

\FloatBarrier
\subsubsection{Localized Predictive-Coding Architectures}\label{localized-predictive-coding-architectures}

The decoupled dual framework can be implemented as a layer-wise predictive-coding system. The forward branch retains the hierarchical prediction chain, while a backward branch supplies a reference close to each forward representation. Instead of assigning every parameter through one terminal loss, layer \(i\) forms the intrinsic discrepancy

\begin{equation}
\boldsymbol{\varepsilon}^{[i]}=\mathbf{y}_{*}^{[i]}-\mathbf{y}^{[i]}.
\end{equation}

The local error module and the parameter block that it regulates therefore share the same layer boundary. This organization preserves the forward representation learned by the deep network while changing the location at which online correction signals are generated.

\paragraph{Guidance-backpropagation architecture.}

\begin{figure}[h]
	\begin{center}\footnotesize
		\begin{tabular}{c}
			\includegraphics[width=.82\textwidth]{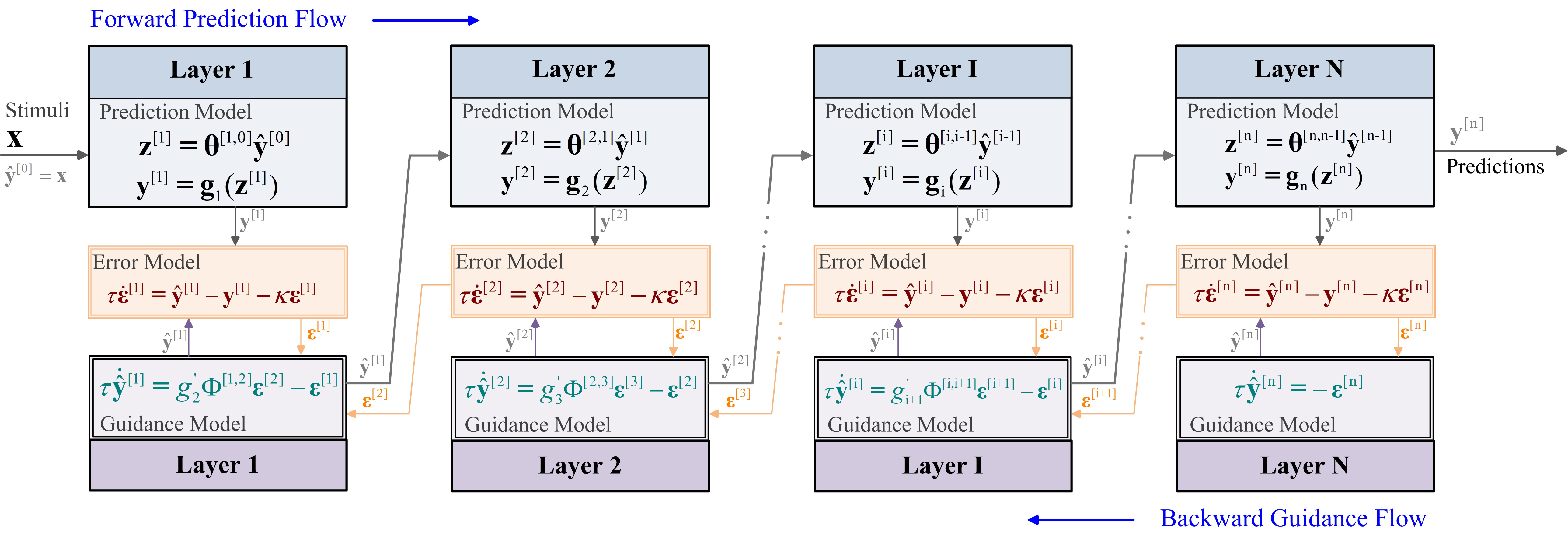}\\[0.7ex]
			a) Predictive Coding Structured Guidance Flow\\[1.4ex]
			\includegraphics[width=.82\textwidth]{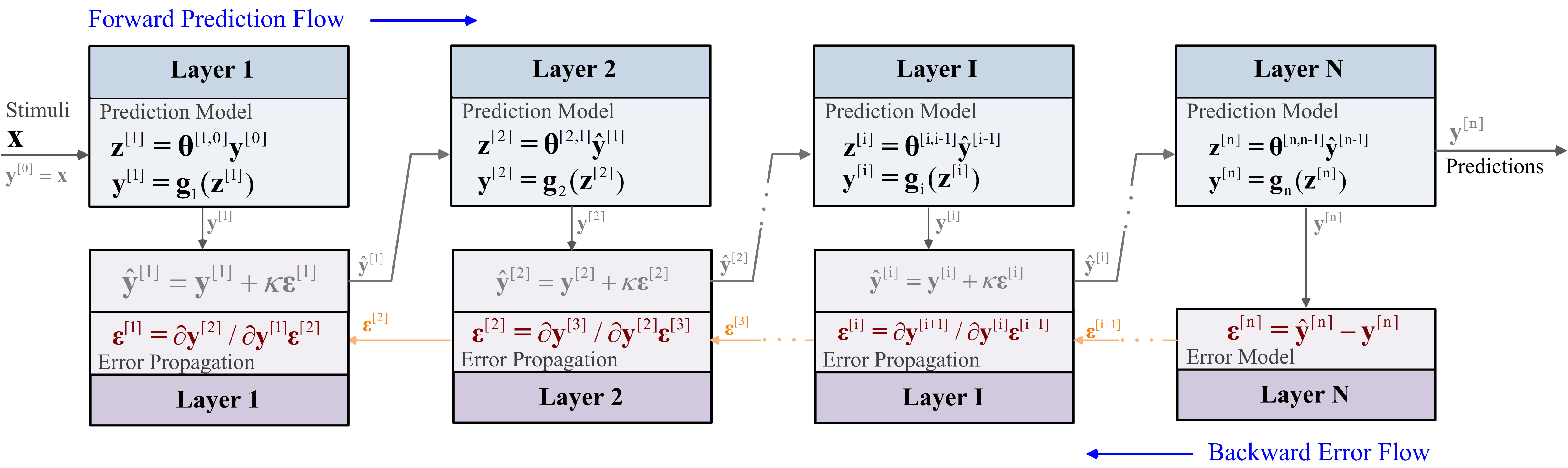}\\[0.7ex]
			b) Predictive Coding Structured Quasi-Steady-State Error Flow\\[1.4ex]
			\includegraphics[width=.82\textwidth]{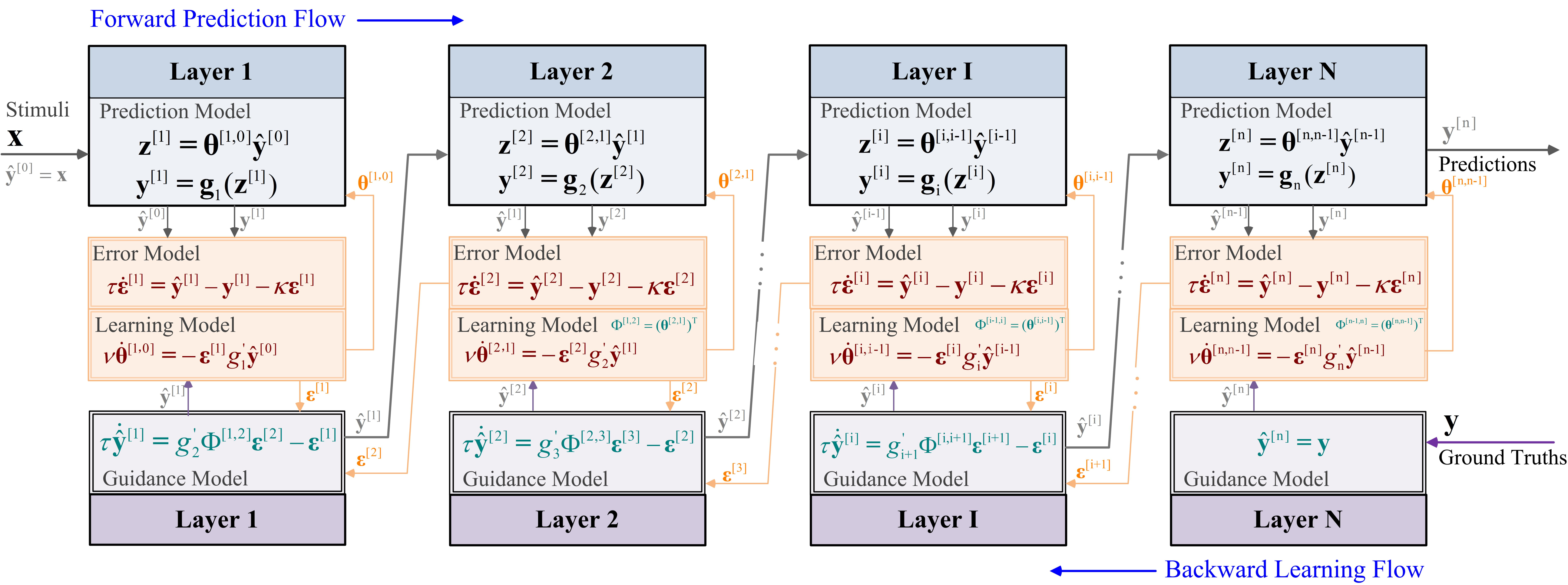}\\[0.7ex]
			c) Predictive Coding Structured Learning Flow
		\end{tabular}
	\end{center}
	\vspace{0.3em}
	\caption{Predictive Coding (Guidance Backpropagation) Architectures}
	\label{fig:guidance-pc}
\end{figure}

In Fig.~\ref{fig:guidance-pc}, the backward guidance flow propagates adjacent-layer error information through the guidance models. At each layer, the generated reconstruction \(\hat{\mathbf{y}}^{[i]}\) is compared with the forward representation \(\mathbf{y}^{[i]}\), producing a localized discrepancy for the corresponding learning block. Panel (a) shows the guidance flow, panel (b) the quasi-steady-state error flow, and panel (c) the resulting structured learning flow.

\FloatBarrier
\FloatBarrier

\begin{figure}[h]
\begin{center}\footnotesize
\begin{tabular}{c}
\includegraphics[width=.82\textwidth]{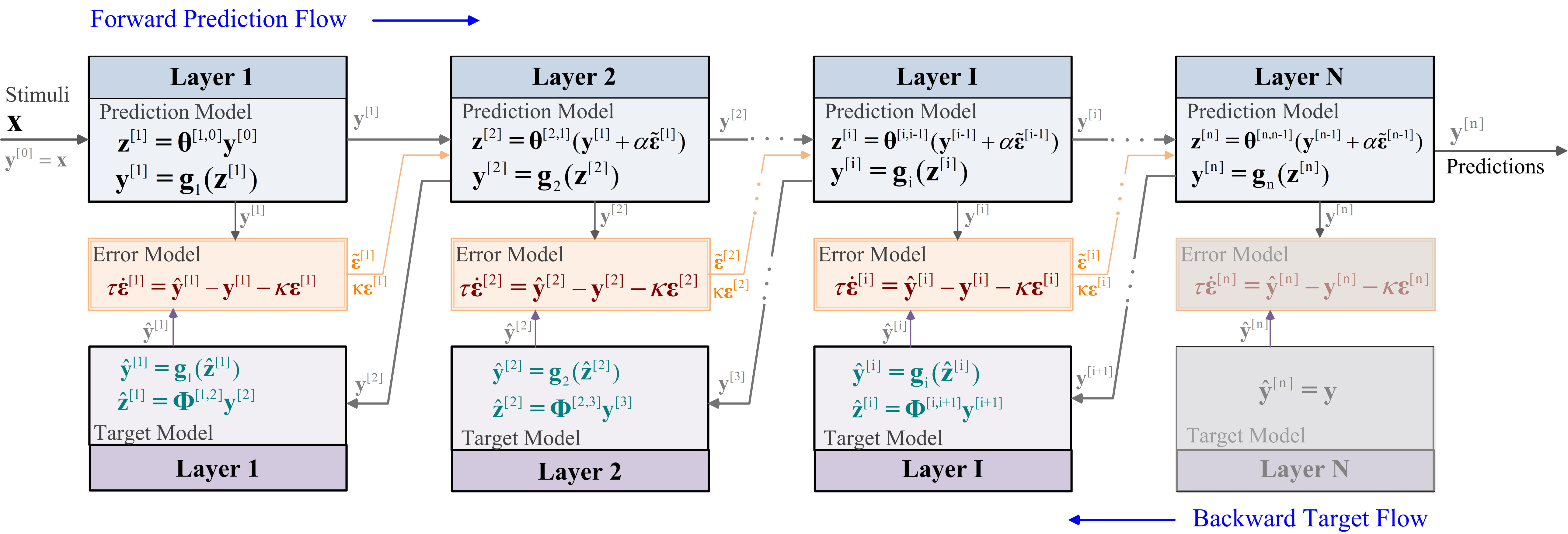}\\[0.7ex]
a) Predictive Coding Structured Transit Target Flow\\[1.4ex]
\includegraphics[width=.82\textwidth]{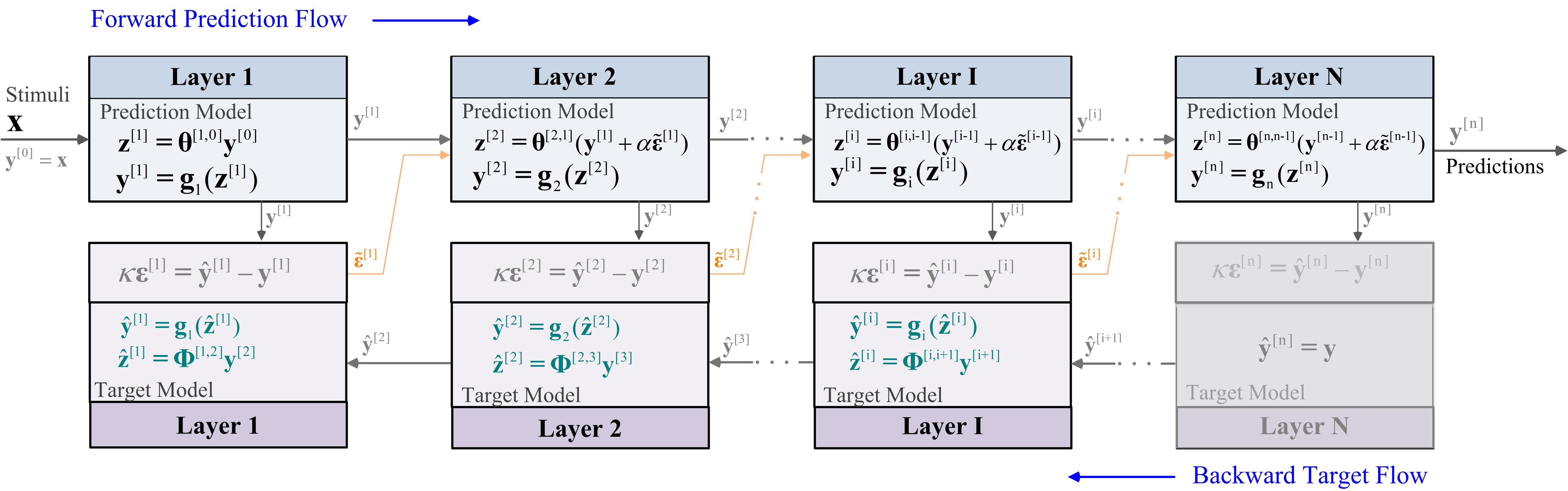}\\[0.7ex]
b) Predictive Coding Structured Quasi-Steady-State Target Flow\\[1.4ex]
\includegraphics[width=.82\textwidth]{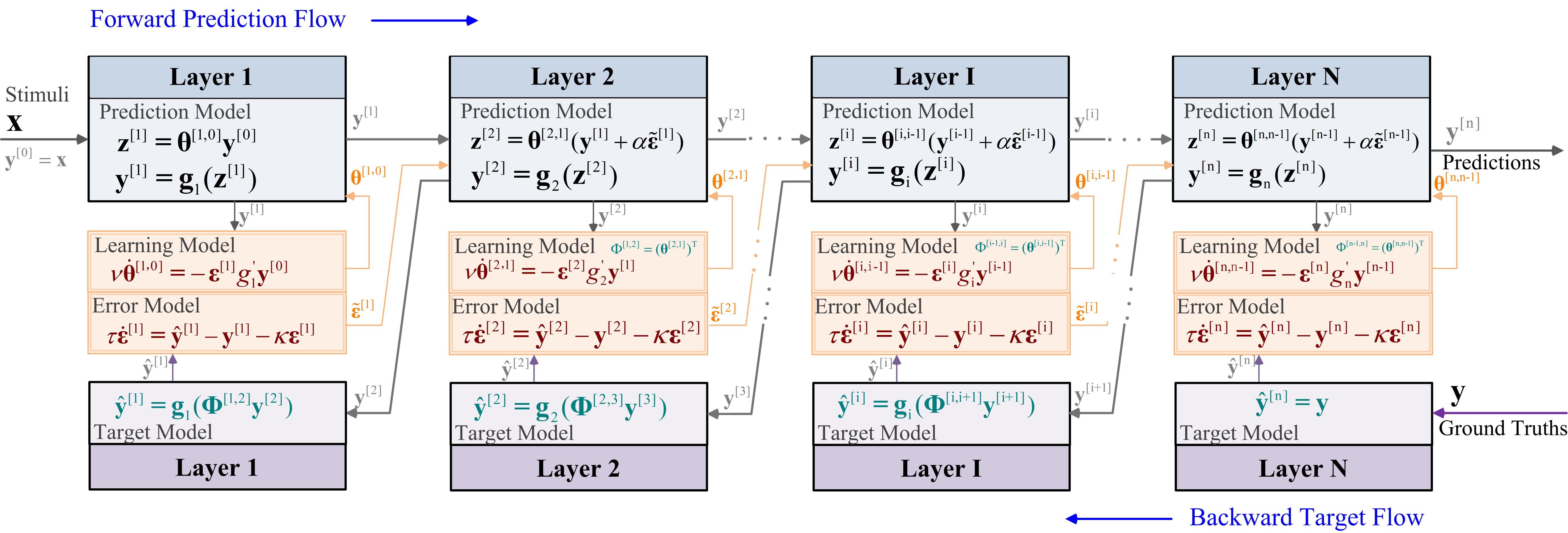}\\[0.7ex]
c) Predictive Coding Structured Learning Flow
\end{tabular}
\end{center}
\vspace{0.3em}
\caption{Predictive Coding (Target Backpropagation) Architectures}
\label{fig:target-pc}
\end{figure}

\paragraph{Target-backpropagation architecture.}
Fig.~\ref{fig:target-pc} uses a generative target path. Beginning with the terminal reference, the backward target model propagates layer-wise reference signals from \(\mathbf{y}_{*}^{[i+1]}\) to \(\mathbf{y}_{*}^{[i]}\) according to

\begin{equation}
\mathbf{y}_{*}^{[i]}=f_{\mathbb{NN},i+1}^{*}
\left(\mathbf{y}_{*}^{[i+1]},\bm{\Phi}^{[i+1]}\right),
\end{equation}

where the backward parameter matrix is generally defined as the transpose of the corresponding forward parameter matrix:

\begin{equation}
\bm{\Phi}^{[i+1]}=\left(\bm{\uptheta}^{[i+1]}\right)^{T}.
\end{equation}

Each generated target is compared with the forward activation at the same layer. Panel (a) displays the transit target flow, panel (b) its quasi-steady routing, and panel (c) the resulting local learning architecture. This realization transmits a sequence of representational targets rather than a single end-to-end gradient.

\FloatBarrier

The two architectures differ in the content carried by the backward branch. Guidance backpropagation propagates adjacent-layer error information to form a local reconstruction, whereas target backpropagation generates an explicit layer-wise target from the higher-level target stream. Both retain bottom-up prediction, form discrepancies locally, and permit layer-wise correction without routing every online update through a single terminal residual. They provide explicit engineering implementations of localized predictive coding. Their complete high-dimensional stability conditions are developed in the companion stability paper.

\FloatBarrier
\subsubsection{Hybrid and Scaled Frameworks}\label{hybrid-and-scaled-frameworks}

High-dimensional environments and complex interactive tasks require sequential and parallel expansion of the prediction, motion, and reconstruction components. The scaled framework in Fig.~\ref{fig:practical} enlarges internal capacity while retaining localized learning. The motion module represents temporal transitions, the prediction branch generates environmental projections, and the reconstruction branch enforces measurement tracking. Sparse localized updates isolate most layer dynamics, while dense local learning can be retained where maximum representational fidelity is required.

\begin{figure}[h]
	\centering
	\includegraphics[width=\textwidth]{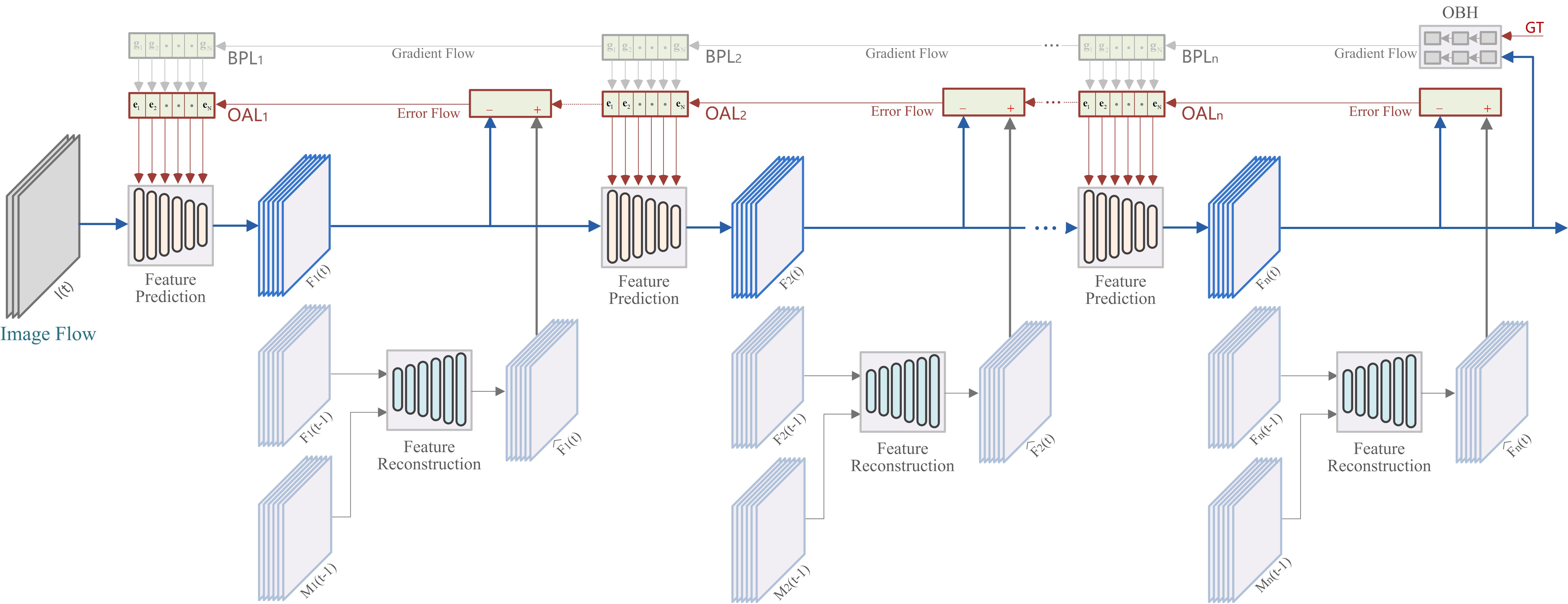}
	\caption{Hybrid Practical Active Learning Framework}\label{fig:hybrid}
\end{figure}

The hybrid configuration in Fig.~\ref{fig:hybrid} combines block-wise backpropagation loops for offline pretraining with decentralized OAL loops for deployment. Ground-truth supervision through an object-detection head may be used during pretraining, while online residuals govern real-time adaptation. The architecture therefore does not discard offline learning; it confines it to the tasks and stages where labels are available and preserves a separate local mechanism for continued adaptation.

\begin{figure}[h]
	\centering
	\includegraphics[width=\textwidth]{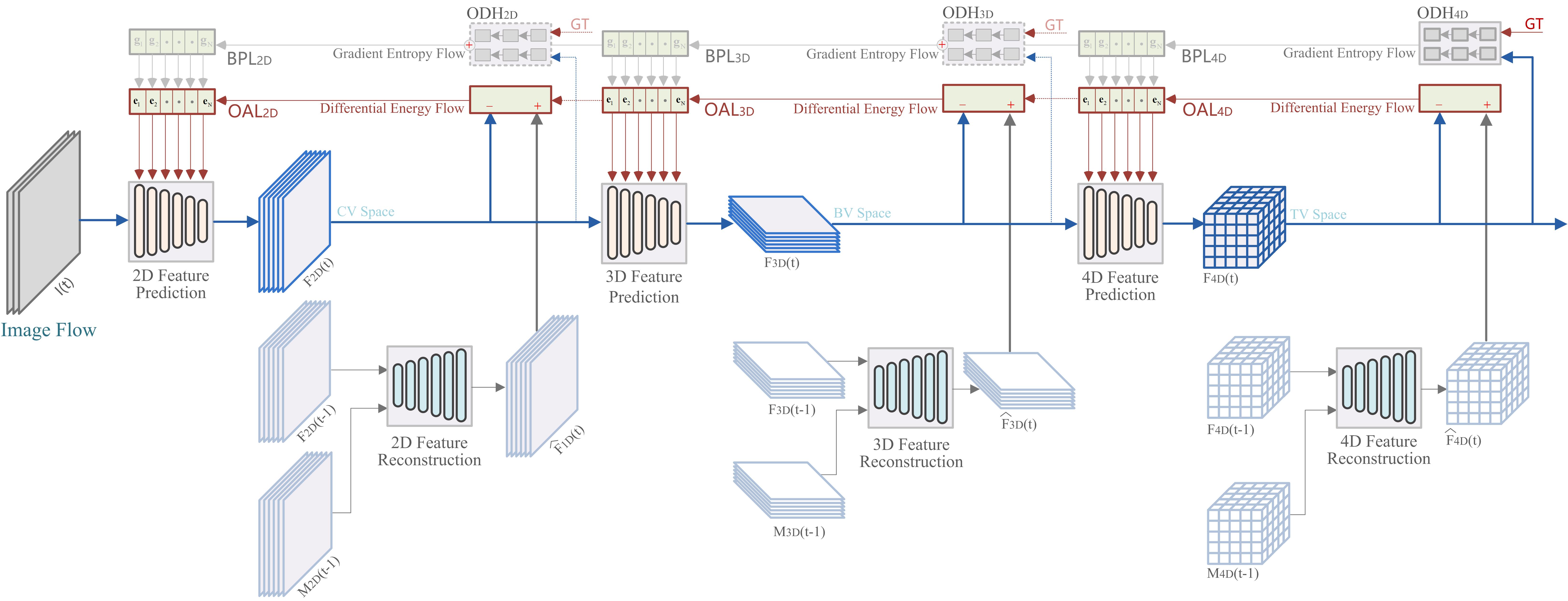}
	\caption{Spatial and Temporal Fused Hybrid Active Learning Framework}\label{fig:fused}
\end{figure}

The spatial and temporal fused hybrid framework extends the same design across two-dimensional planar, three-dimensional spatial, and four-dimensional spatiotemporal representations, as shown in Fig.~\ref{fig:fused}.

Sequential expansion increases the depth of the represented dynamics, while parallel expansion accommodates multiple channels or subsystems. Conventional end-to-end backpropagation is not required to carry every online error through the full topology. Local differential-energy and gradient-information flows are routed across decoupled blocks instead. The network can therefore modify its internal representation in a non-stationary tracking environment without forcing all layers to depend on one terminal residual. This decentralized organization provides the architectural basis for scaling online adaptation to complex multi-step environments.

\FloatBarrier
\section{Conclusion}\label{conclusion}

This paper establishes an Online Active Learning (OAL) paradigm for Control AI. The motivation follows from a structural mismatch: offline deep learning supplies high-capacity hierarchical representation but normally freezes learning during deployment, whereas adaptive learning supplies continuous error correction but becomes difficult to scale when its internal dynamics are densely cross-coupled. The proposed paradigm combines the two by embedding a structured neural model in a continuous Prediction-Evaluation-Correction loop.

The principal architecture separates prediction, latent dynamics, and reconstruction. The primary framework moves the loop to the sensory boundary and uses the reconstruction residual as an intrinsic learning signal. Spatial cross-reconstruction and temporal displacement prevent the system from solving the task through a trivial identity map. Coupled and dual configurations establish corresponding prediction and reconstruction layers. Localized predictive coding then implements layer-wise correction through either reconstruction-based guidance backpropagation or generative target backpropagation. These local residuals shorten error paths, while hybrid and scaled architectures connect offline pretraining with continued real-time adaptation across larger spatial and temporal representations.

The present paper therefore contributes a unified online-learning paradigm rather than a task-specific model: it defines the architecture, the roles of its prediction and reconstruction processes, and the relationships among its global and local learning loops. It retains the local Lyapunov-energy direction needed to explain the update laws, while the complete high-dimensional stability conditions and formal proof are developed in a separate companion paper. Task-specific implementation and experimental validation are beyond the scope of the present architectural study. Within this scope, the paradigm supplies a common basis for continually adaptive perception and control in non-stationary physical environments.

\begingroup
\footnotesize
\setlength{\bibsep}{0pt}
\bibliography{references}
\endgroup
\end{document}